\documentclass[letterpaper, 10 pt, conference]{ieeeconf}  % Comment this line out if you need a4paper

\IEEEoverridecommandlockouts                              % This command is only needed if 
\usepackage{svg}
\usepackage{multicol}
\usepackage{multirow}
\usepackage{tabularx}
\usepackage{booktabs}
\usepackage{graphicx}
\usepackage{float}
\usepackage{orcidlink}
\hypersetup{hidelinks}

\makeatletter
\newcommand{\linebreakand}{%
  \end{@IEEEauthorhalign}
  \hfill\mbox{}\par
  \mbox{}\hfill\begin{@IEEEauthorhalign}
}
\makeatother

\title{\LARGE \bf
3D Understanding of Deformable Linear Objects: \\ Datasets and Transferability Benchmark
}

\begin{document}

\author{Bare Luka Zagar\orcidlink{0000-0001-5026-3368},
Tim Hertel\orcidlink{0009-0002-5162-6401},
Mingyu Liu\orcidlink{0000-0002-8752-7950},
Ekim Yurtsever\orcidlink{0000-0002-3103-6052}, \IEEEmembership{Member, IEEE}, \linebreakand
Alois Knoll\orcidlink{0000-0003-4840-076X}, \IEEEmembership{Fellow, IEEE}
    % <-this % stops a space
\thanks{This result is part of a project that has received funding from the European Union's Horizon 2020 research and innovation programme under grant agreement No 870133.}
\thanks{BL. Zagar, T. Hertel, M. Liu, and AC. Knoll are with the Chair of Robotics, Artiﬁcial Intelligence and Real-Time Systems, Technical University of Munich, 85748 München, Germany (E-mail: bare.luka.zagar@tum.de, tim.hertel@tum.de, mingyu.liu@tum.de, hu.nguyen@tum.de, knoll@in.tum.de)}
\thanks{E. Yurtsever is with the College of Engineering, Center for Automotive Research, The Ohio State University, Columbus, OH 43212, USA (E-mail: yurtsever.2@osu.edu)}
\thanks{This work has been submitted to the IEEE for possible publication. Copyright may be transferred without notice, after which this version may no longer be accessible.}
}

\maketitle

% \twocolumn[{%
% \renewcommand\twocolumn[1][]{#1}%
% \maketitle
% \begin{center}
%     \centering
%     \includegraphics[width=0.95\textwidth]{ieeeconf/figures/drawing-ral.png}
%     %\vspace{-5mm}
%     \captionof{figure}{Deformable linear objects can manifest in many variations, shapes, and configurations. For example,  automotive wiring harnesses (\textbf{left}) and  blood vascular structures (\textbf{right}) are completely different types of objects, but when compared at a micro-level, it is clear that they share similar spatial topological properties.}
%     \label{fig:graph_abstract}
% \end{center}%
% }]

% \maketitle
% \begin{figure*}
%     \centering
%     \includegraphics[width=0.95\textwidth]{ieeeconf/figures/figure_1_v3.jpg}
%     \caption{Deformable linear objects can manifest in many variations, shapes, and configurations. For example,  automotive wiring harnesses (\textbf{left}) and  blood vascular structures (\textbf{right}) are completely different types of objects, but when compared at a micro-level, it is clear that they share similar spatial topological properties.}
%     \label{fig:graph_abstract}
% \end{figure*}
% \thispagestyle{empty}
% \pagestyle{empty}
%%%%%%%%%%%%%%%%%%%%%%%%%%%%%%%%%%%%%%%%%%%%%%%%%%%%%%%%%%%%%%%%%%%%%%%%%%%%%%%%
\begin{abstract}
Deformable linear objects are vastly represented in our everyday lives. It is often challenging even for humans to visually understand them, as the same object can be entangled so that it appears completely different. Examples of deformable linear objects include blood vessels and wiring harnesses, vital to the functioning of their corresponding systems, such as the human body and a vehicle. However, no point cloud datasets exist for studying 3D deformable linear objects. Therefore, we are introducing two point cloud datasets, PointWire and PointVessel.  
We evaluated state-of-the-art methods on the proposed large-scale 3D deformable linear object benchmarks. Finally, we analyzed the generalization capabilities of these methods by conducting transferability experiments on the PointWire and PointVessel datasets.
\end{abstract}

%%%%%%%%%%%%%%%%%%%%%%%%%%%%%%%%%%%%%%%%%%%%%%%%%%%%%%%%%%%%%%%%%%%%%%%%%%%%%%%%
\section{Introduction}

\begin{figure}[t!]
    \centering
    \includegraphics[width=0.95\columnwidth]{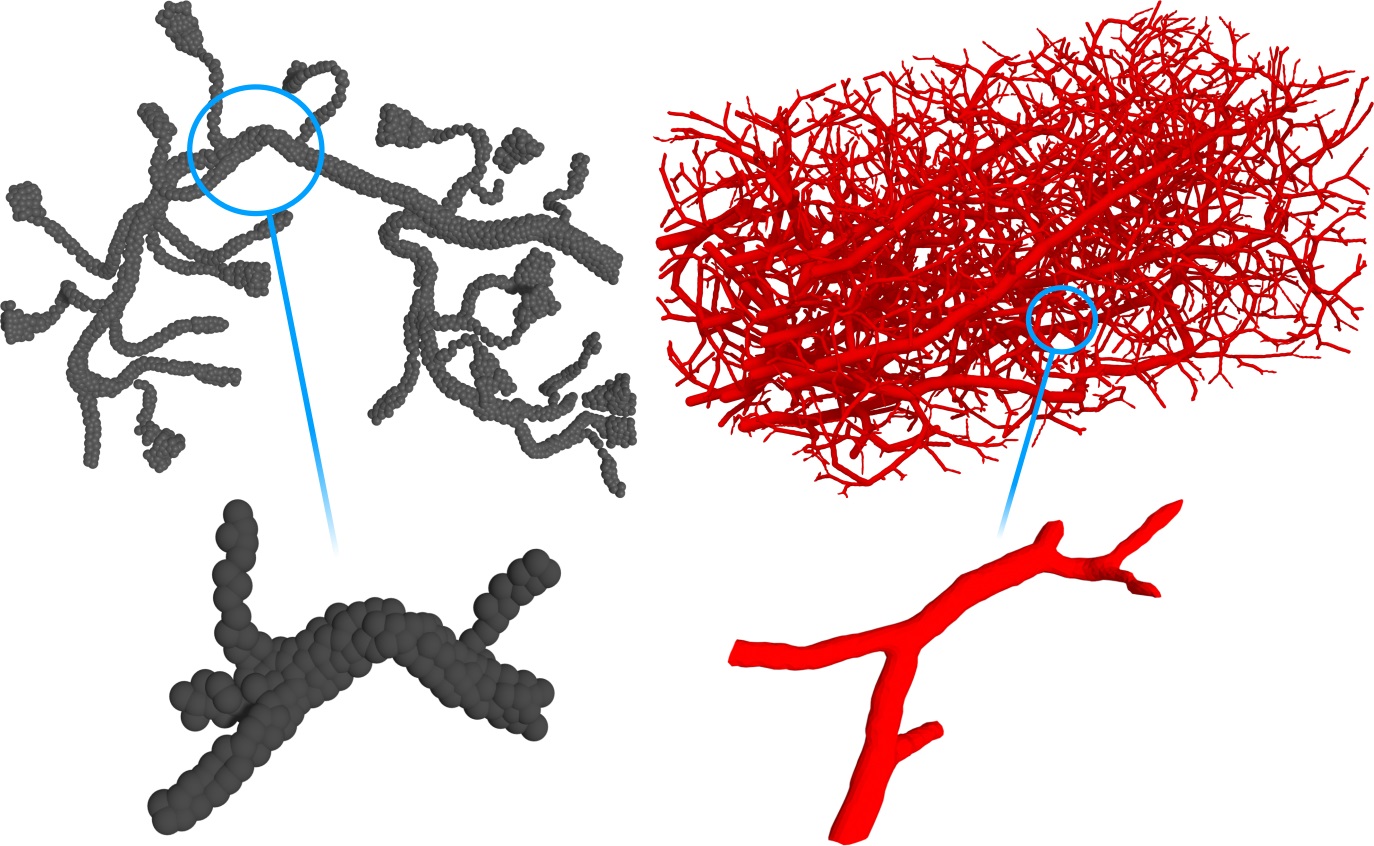}
    \caption{Deformable linear objects can manifest in many variations, shapes, and configurations. For example,  automotive wiring harnesses (\textbf{left}) and blood vascular structures (\textbf{right}) are completely different types of objects, but when compared at a micro-level, it is clear that they share similar spatial topological properties.}
    \vspace{-5mm}
    \label{fig:graph_abstract}
\end{figure}

Deformable linear objects (DLOs), such as blood vessels and wiring harnesses, are crucial components of their corresponding higher-level systems. Blood vessels allow the vascular system to carry blood throughout the human body, while wiring harnesses enable the electrical and electronic architecture to transport current through the vehicle body. Despite their inherent differences, when observed through the prism of DLOs, these objects are based on the same underlying topological structure. Moreover, DLOs can come in very complex entangled configurations, as shown in Fig.~\ref{fig:graph_abstract}. Therefore, understanding the topology and disentanglement of these structures is a challenging perception task.

The industry of wiring harnesses is very underdeveloped in terms of automation, or to be more precise, the level of automation is close to zero \cite{nguyen2021manufacturing, navas2022wire}. Very little research has been conducted regarding the perception of wiring harnesses. In \cite{nguyen2021novel}, the focus was on extracting the profile of simpler wiring harnesses, while \cite{kicki2021tell} presented an approach to classify different parts of complex automotive wiring harnesses. A sophisticated 2D instance segmentation approach of wires was introduced in \cite{caporali2022fastdlo}. However, the main problem still remains. There is no available 3D dataset of complex wiring harnesses and no studies on extracting topological elements, such as bifurcations and endpoints in order to achieve disentanglement.

On the other hand, deep learning methods have boosted the medical imaging field and are gaining more and more attention \cite{valliani2019deep}. A significant amount of research in the medical imaging domain focuses on the segmentation of vascular systems, such as brain vasculature \cite{todorov2020machine, kirst2020mapping} and retinal vessel segmentation \cite{khandouzi2022retinal}. Recently, the IntrA \cite{yang2020intra} dataset sparked research on point cloud segmentation of intracranial aneurysms \cite{liu2022edge, yang2023two}. However, there still does not exist a publicly available 3D point cloud dataset for studying blood vessels' spatial topology configurations.
%such as understanding the vascular system, there are still some open issues to be tackled, such as the analysis regarding coronary vascular diseases\cite{hampe2019machine} and precise drug delivery \cite{liu2019precise}.

Therefore, we propose two deformable linear object datasets, PointWire and PointVessel. The PointWire dataset is based on 40 real-scanned complex automotive wiring harnesses, which we extended with our semi-automatic dataset generator to a total of $12000$ samples. The high-precision raw point clouds, including a wiring harness, are first processed by removing the background and noise. Then, a skeleton of the wiring harness points is retrieved and manually refined. The refined skeleton is used for automating the process of rigging the wiring harness point cloud with Blender \cite{blender}. It enables us to extend the dataset size and introduce a wide variety of wiring harness spatial configurations. The PointVessel dataset is derived from \cite{dvn}, which includes 136 blood vessel volumes. Firstly, we convert the volumes to mesh data and separate each volume into 96 cropped-out smaller vascular structures. Then, the surface of these crops is uniformly subsampled to retrieve the final point cloud data.

There are multiple benefits of the introduced PointWire and PointVessel datasets: 1) more research regarding the automation of wiring harnesses will be encouraged, 2) the different data domains but still very similar data structure and topology will further boost the research in the medical imaging field of vascular systems by providing a different perspective on similar underlying issues, and 3) due to a general lack of data, the manufacturing and medical imaging fields can benefit from jointly using the given datasets in terms of transfer learning.

The key contributions of our work are the following:
\begin{itemize}
    \item We are the first to introduce point cloud datasets - PointWire and PointVessel - for understanding DLOs by means of disentanglement and segmentation of topological structures.
    \item We propose a semi-automatic pipeline for processing, augmenting, and annotating real-scanned wiring harnesses.
    \item We provide large-scale benchmarks and transferability analysis for understating the typology of the complex DLOs  by combining segmentation and disentanglement. 

\end{itemize}

%------------------------------------------------------------------------
\section{Related Work}

%------------------------------------------------------------------------
\subsection{3D Object Datasets}
In the last decade, 3D datasets enabled the tremendous advancement of deep learning in the point cloud domain. 
%Numerous 3D LiDAR-based datasets \cite{kitti, caesar2020nuscenes, waymo, chang2019argoverse} accelerated the autonomous driving domain.
Many 3D object datasets, that employ synthetic CAD models, are still actively used for common tasks such as point cloud classification and segmentation. The most popular ones are the ModelNet \cite{modelnet} and ShapeNet \cite{chang2015shapenet} datasets that contain a different amount of object types: cars, planes, guitars, chairs, tables, and many more. ABO \cite{collins2022abo} presented a more realistic and real-world relevant 3D object dataset. Additionally, the ABC \cite{koch2019abc} and DeepCAD \cite{wu2021deepcad} introduced a dataset based on mechanical objects. The major problem of the synthetic CAD model-based datasets is the lack of realistic propertiers which renders them useless for real-world usage. Therefore, ScanObjectNN \cite{uy-scanobjectnn-iccv19} and OmniObject3D \cite{wu2023omniobject3d} proposed real-scanned 3D objects for the creation of their datasets. However, most of the aforementioned datasets contain only common objects. Recently, \cite{yang2020intra} introduced the IntrA dataset to encourage the research on point cloud segmentation of aneurysms. Unfortunately, the IntrA dataset is relatively small and does not focus on the complexity of extracting the blood vessel topology.

Our proposed PointWire and PointVessel datasets enable the research of segmenting the topological features of complex deformable linear objects, such as wiring harnesses and blood vessels.

%------------------------------------------------------------------------
\begin{figure*}[t!]
    \centering
    \includegraphics[width=0.99\textwidth]{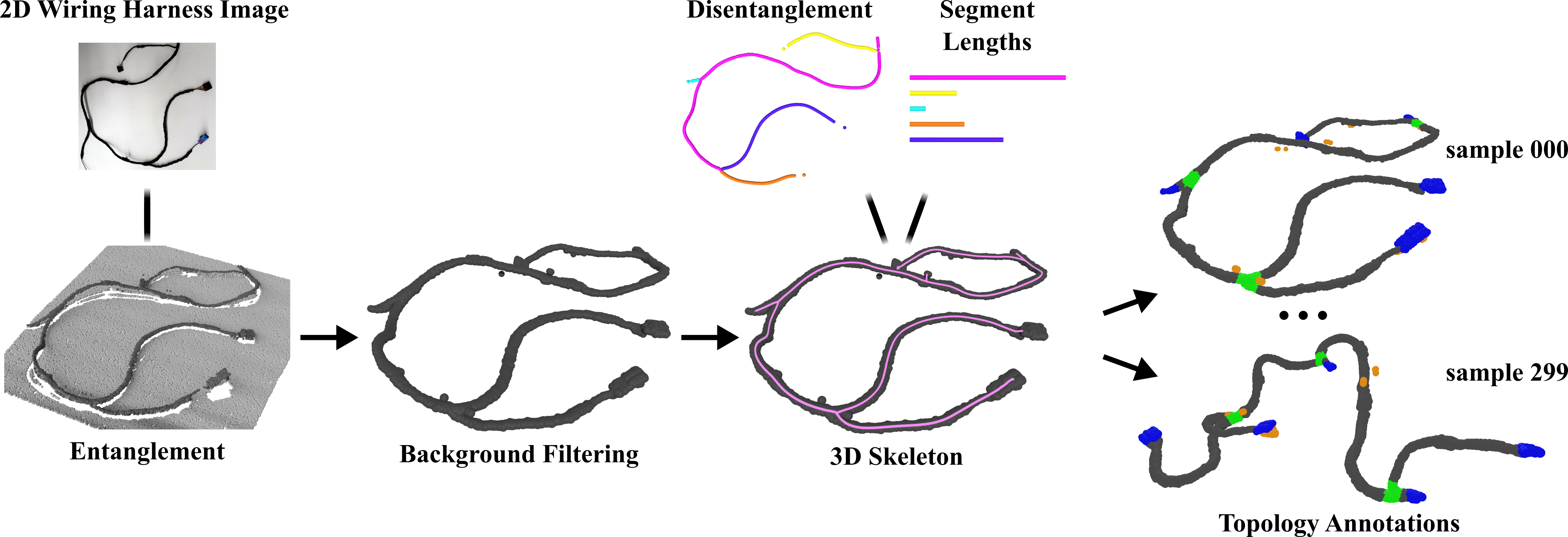}
    \caption{Wiring harness dataset generation pipeline: First, the raw input point cloud of an entangled wiring harness is taken with a Photoneo MotionCam-3D M. Then, we perform plane segmentation and statistical noise removal. Afterward, a skeleton is retrieved using the Laplace-based contraction method \cite{laplace}. Additionally, the skeleton is transformed into a disentangled DLO graph representation. Further, the clean wire harness point cloud and the skeleton-based graph are combined and extended using the Blender toolkit. Finally, a diverse set of wire harness configurations is achieved.}
    \label{fig:wh_pipeline}
\end{figure*}

\subsection{Point Cloud Segmentation}
In recent years, point cloud segmentation has been dramatically developed into an indispensable technology fro computer vision. PointNet\cite{qi2017pointnet} exploits multi-layer perceptron network to extract features from point clouds.
% is a pioneering work using multi-layer perceptron network to directly extract point-wise features from the input point clouds. 
PointNet++\cite{qi2017pointnet++} utilizes a local region grouping module and hierarchical PointNet to capture regional information. \cite{li2018pointcnn, xu2018spidercnn, lin2020fpconv, xu2021paconv} explore using convolution neural network to learn point features. 
% Point2Sequence\cite{liu2019point2sequence} introduce attention mechanism to aggregate contextual information of local regions. 
% Graph-based approaches \cite{wang2019dynamic, wang2018local, wang2019graph, xu2020grid, zhou2021adaptive, hui2022learning} exploit the graph neural network to gather geometric information from point clouds. 
DGCNN\cite{wang2019dynamic} proposes using the EdgeConv module to obtain edge features from the KNN-based local graphs. 
% AdaptConv\cite{zhou2021adaptive} generates an adaptive kernel for points according to their dynamically learned features. 
With the popularity of Transformer, PointTransformer\cite{zhao2021point} and PCT\cite{guo2021pct} explore transformer-based networks and achieve state-of-the-art performance. 
% Methods\cite{ran2021learning, yan2020pointasnl, xiang2021walk} design local aggregators via relations of points. 
% RPNet\cite{ran2021learning} computes features based on the geometric and semantic relations of point clouds. 
CurveNet\cite{xiang2021walk} implements a curve grouping operator to aggregate point features. The most recent work RepSurf\cite{ran2022surface} exploits triangle-based and multi-surface representation for segmentation. DeltaConv\cite{Wiersma2022DeltaConv} proposes a set of anisotropic convolution layers for point cloud surface representation. 

Although these aforementioned methods have been widely utilized in common object point cloud segmentation, their performances have not been verified on deformable linear objects due to a lack of related datasets. Moreover, there is a lack of systematic analysis of the model’s ability to understand general concepts within the scope of point cloud segmentation. Hence, to mitigate these shortcomings, we evaluate several state-of-the-art methods on our large-scale benchmarks. 

\subsection{Deformable Linear Objects Perception}
DLOs in the form of blood vessels are intensively studied in the medical imaging field \cite{mookiah2021review, fu2020rapid}. DeepVesselNet \cite{dvn} introduced a synthetically generated blood vessel dataset and proposed a 3D U-Net architecture-like solution to retrieve the topological information of blood vessels, such as bifurcations and centerlines. Recently, VesselGraph \cite{paetzold2021whole} collected available blood vessel datasets and generated a large-scale graph dataset. However, no available dataset focuses on the 3D spatial attributes that point cloud data could provide. Therefore, we propose to fill this gap with our PointVessel dataset, which enables the study of blood vessels in the rich 3D spatial space.

Cables, wires, ropes and wiring harnesses are another important type of DLOs, which are unfortunately very understudied in the perception research domain. Only recently, a few researchers focused on real-world 2D image segmentation of wires \cite{caporali2022ariadne+,choi2023mbest, rtdlo}. In \cite{lv2022learning} the authors proposed a solution for robustly estimating the 3D state of DLOs from point cloud data. A synthetic point cloud dataset of complex DLOs, i.e.~wiring harnesses, was presented in \cite{nguyen2022enabling}. Unfortunately, the dataset is not publicly available which makes further research impossible. Moreover, the presented experimental anaylsis and the scale of the dataset in \cite{nguyen2022enabling} are rather low and more study is required. Therefore, we intend to mitigate these drawbacks with our PointWire datasets. The PointWire dataset is a large-scale point cloud DLO dataset based on real-scanned wiring harnesses.

\section{DLO Datasets}

Since there are no available point cloud datasets for complex DLOs, such as wiring harnesses and blood vessels, we are the first to propose datasets of such kind. The PointWire and PointVessel datasets serve for topological DLOs key part segmentation and as a transferability benchmark to study the generalization capabilities of point cloud segmentation methods.

\begin{figure}[t!]
	\centering 
    \resizebox{0.45\textwidth}{!}{
    \setlength{\tabcolsep}{1pt}
    \renewcommand{\arraystretch}{2}
	\begin{tabular}[0.8\columnwidth]{cccc} 
	%%%%%%%%%%%%%%%%%%%%%%%%%%%%%%%%%%%%%%%%%%%%%%%%%%%%%%%%%%%%%%%%%%%%%%%%%%%%%%%%%%%%%%%%%%%%%%%%%%%%%%%%%%%%%%%
	   ~ & Sample 021 & Sample 022 & Sample 029 \\
	%%%%%%%%%%%%%%%%%%%%%%%%%%%%%%%%%%%%%%%%%%%%%%%%%%%%%%%%%%%%%%%%%%%%%%%%%%%%%%%%%%%%%%%%%%%%%%%%%%%%%%%%%%%%%%%
		% GT
		\rotatebox[y=9mm]{90}{Raw} \rotatebox[y=9mm]{90}{Point Cloud} \hspace{0.1mm} &
		\includegraphics[width=.28\linewidth]{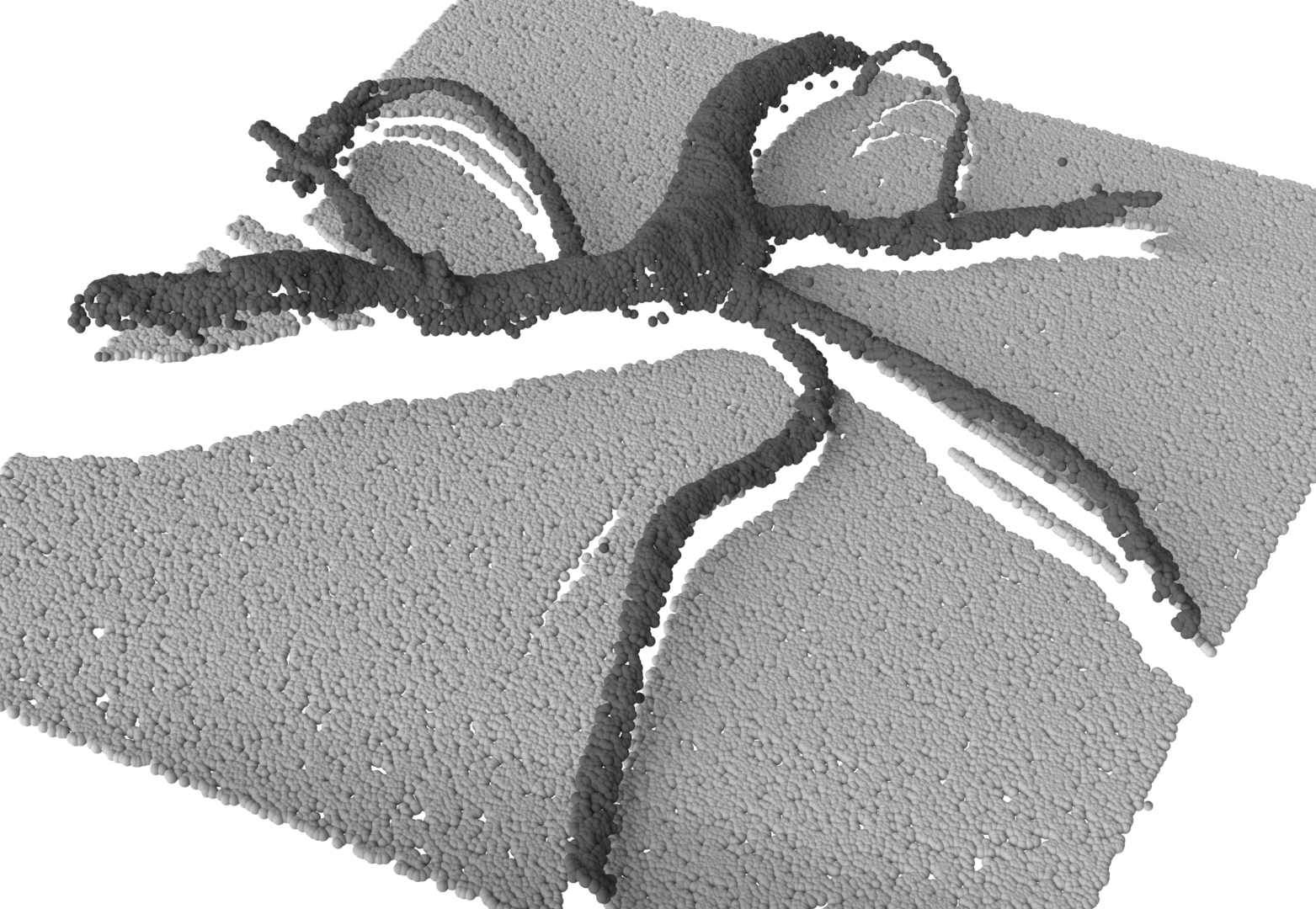} \hspace{-2mm} \vspace{-0.4mm}& 
		\rotatebox0{\includegraphics[width=.28\linewidth]{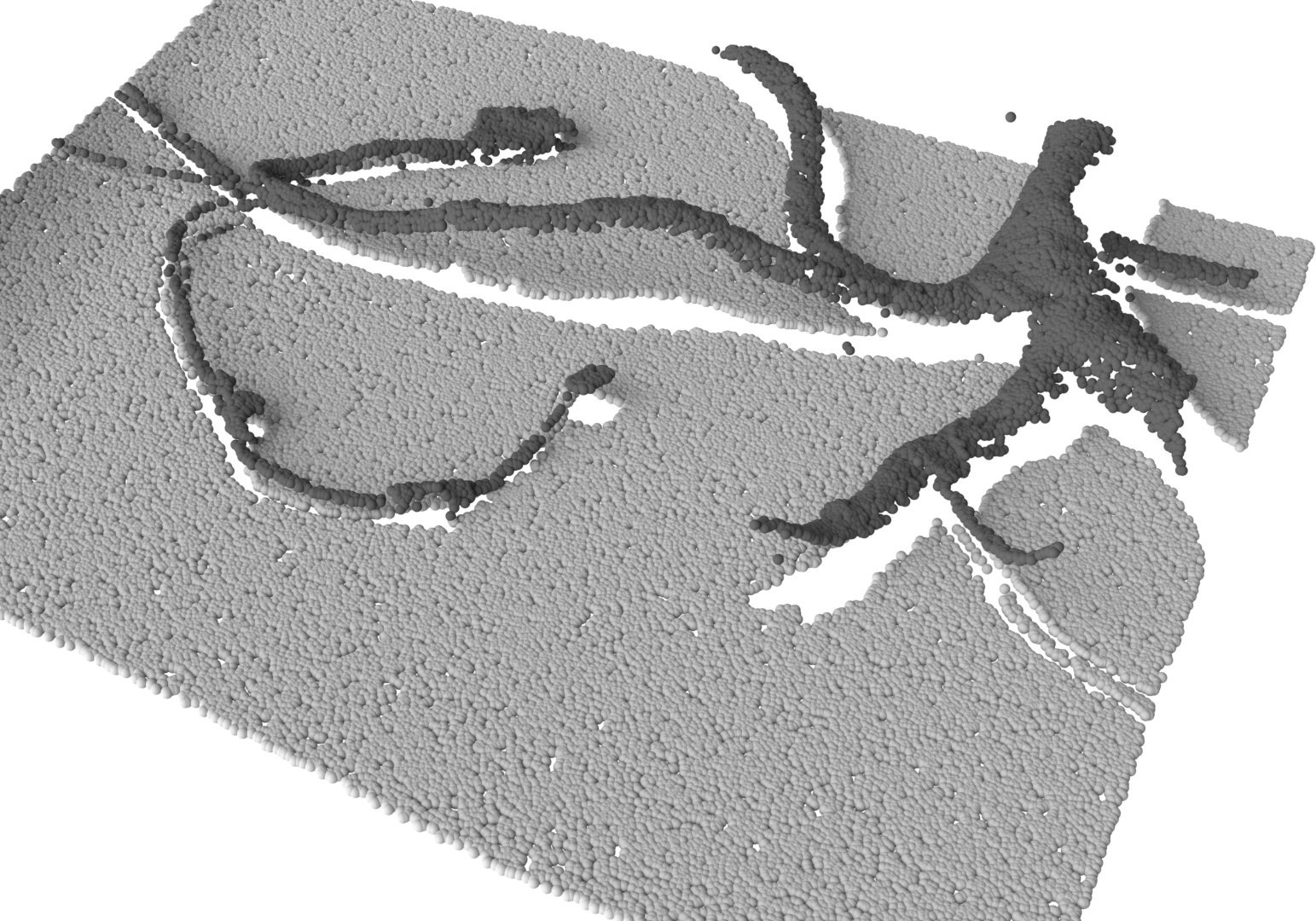}} \hspace{-2mm} \vspace{-0.4mm}&
		\includegraphics[width=.28\linewidth]{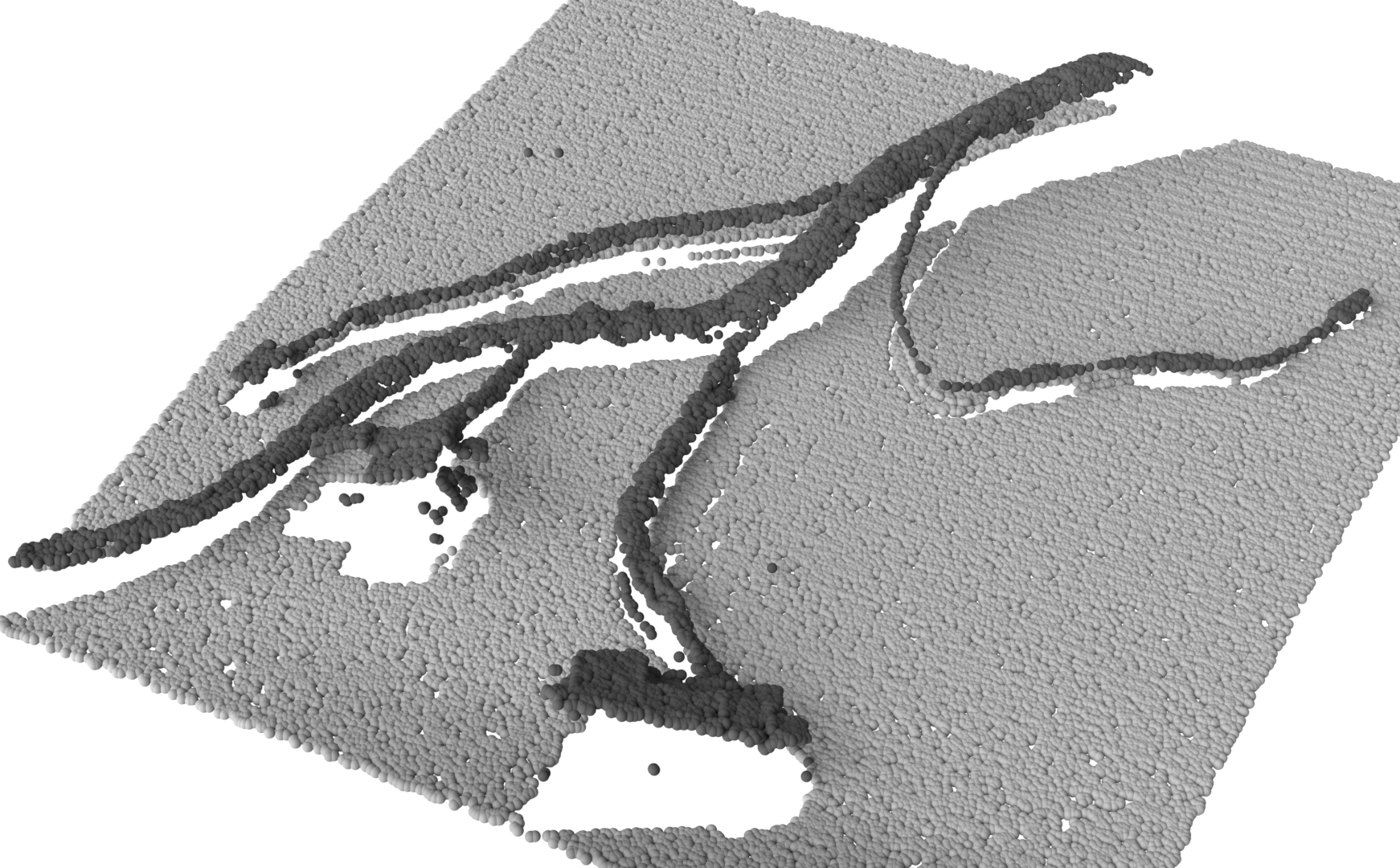} \vspace{-0.4mm}\\
		% Segmentation
		\rotatebox[y=12mm]{90}{Topology} \rotatebox[y=12mm]{90}{Segmentation}\hspace{0.1mm} &
		\includegraphics[width=.28\linewidth]{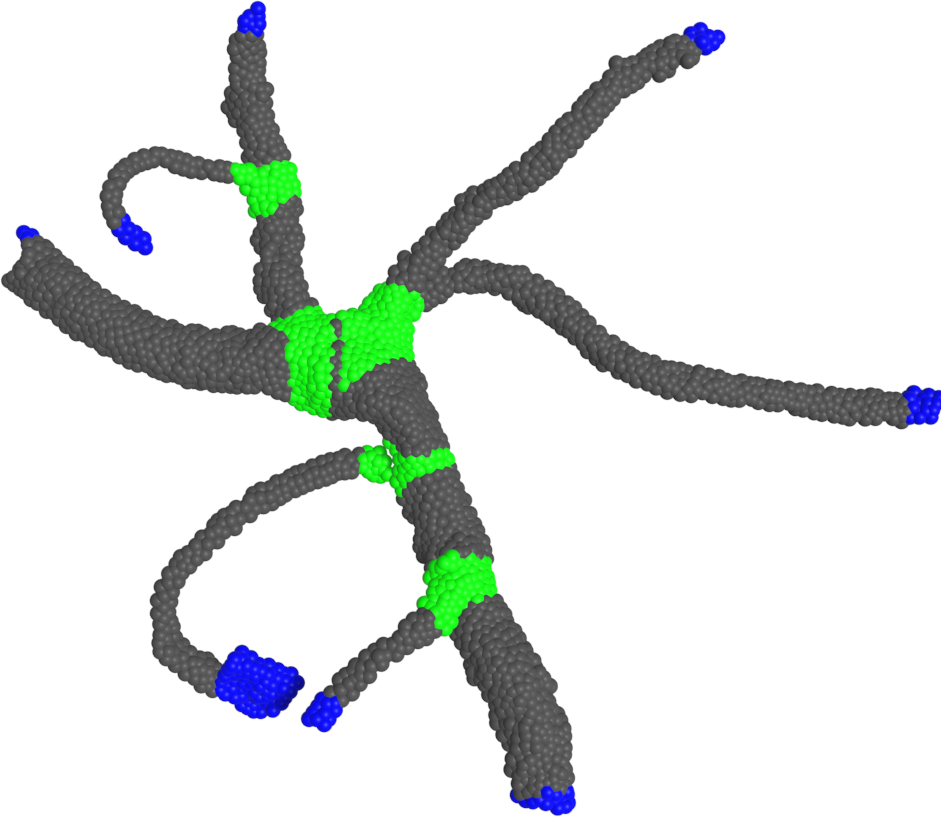} \hspace{-2mm}  \vspace{-1mm} & 
		\rotatebox{45}{\includegraphics[width=.28\linewidth]{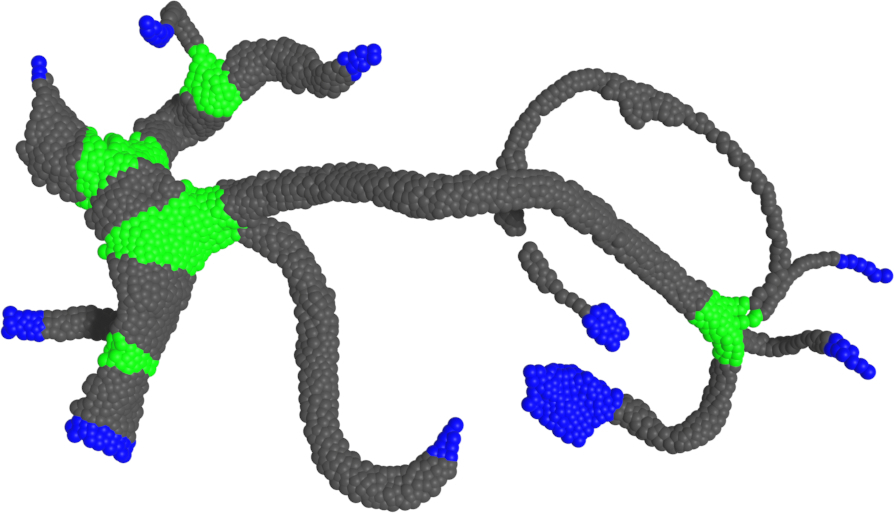}} \hspace{-2mm} \vspace{-1mm} &
		\includegraphics[width=.28\linewidth]{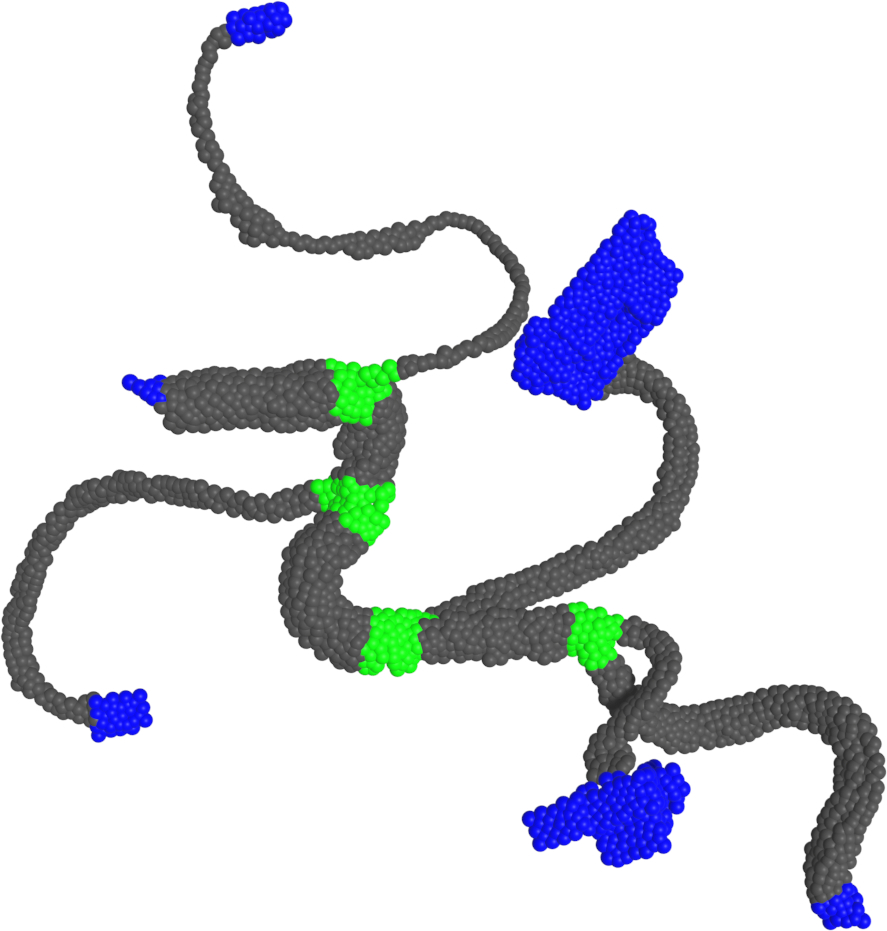} \vspace{-1mm} \\
		% Disentanglement
		\rotatebox[y=12mm]{90}{Disentanglement} \hspace{0.1mm} &
		\includegraphics[width=.28\linewidth]{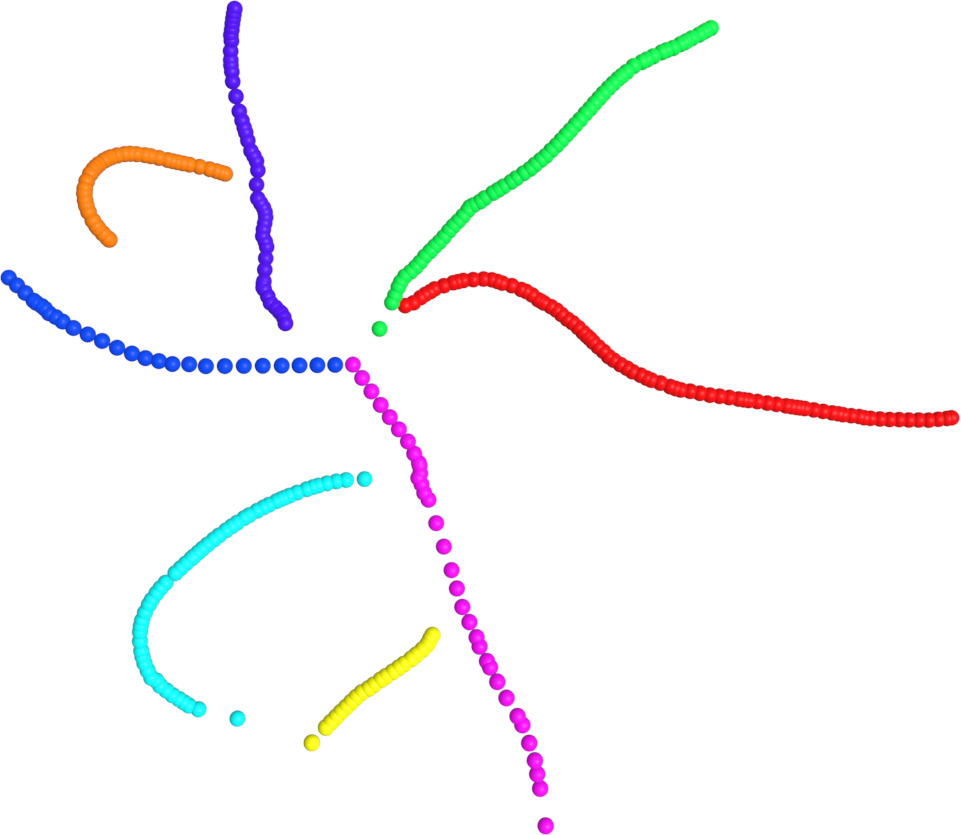} \hspace{-2mm} \vspace{1mm} & 
		\rotatebox{45}{\includegraphics[width=.28\linewidth]{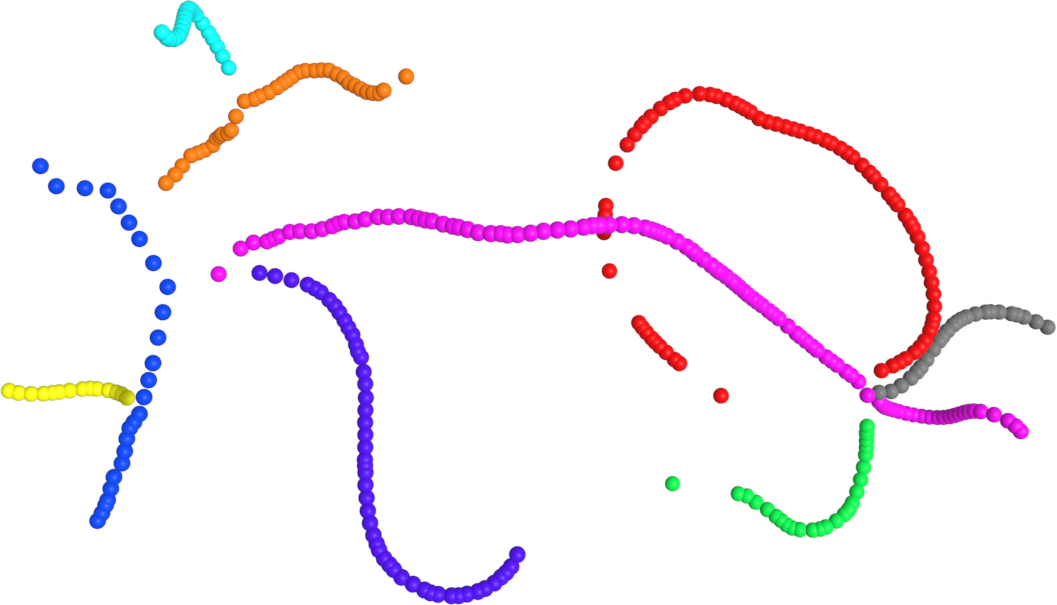}} \hspace{-2mm} \vspace{1mm} &
		\includegraphics[width=.28\linewidth]{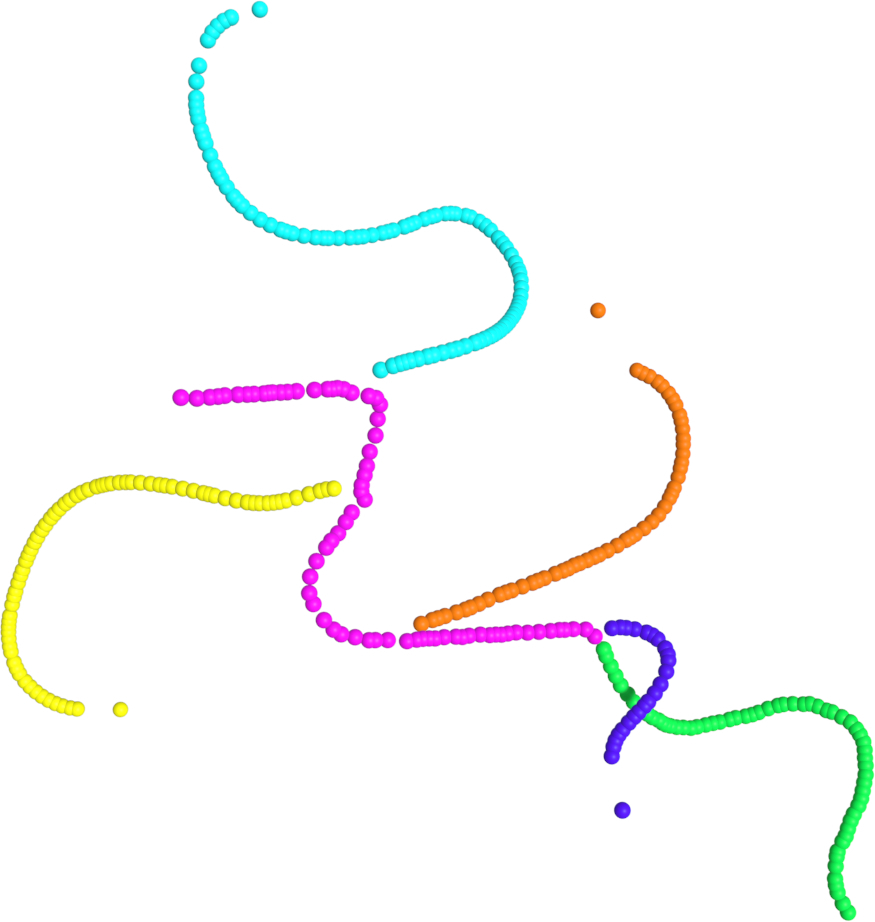} \hspace{0.4mm} \vspace{1mm} \\
% 		% Length
		\rotatebox[y=8mm]{90}{Segment} \rotatebox[y=8mm]{90}{Lengths} \hspace{0.1mm} &
		\includegraphics[width=.22\linewidth]{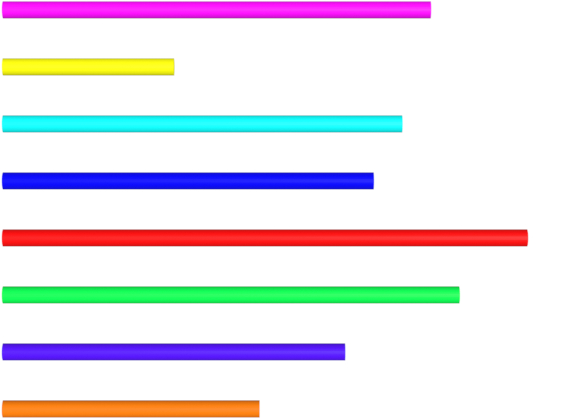} \hspace{-2mm} & 
		\includegraphics[width=.22\linewidth]{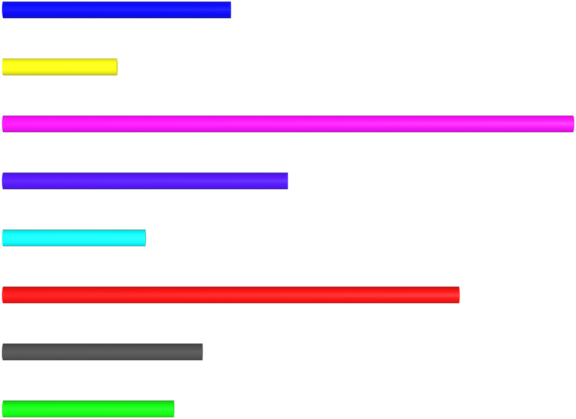} \hspace{-2mm} &
		\includegraphics[width=.22\linewidth]{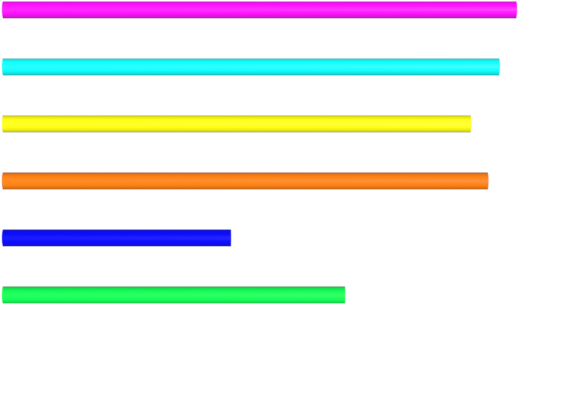} \\
		\hspace{-2mm} 
    
		%\hspace{-0.4mm}
    \end{tabular} 
    }
    \vspace{-10mm}
	\caption{PointWire dataset description: The \textbf{first} row shows entangled wiring harness point clouds. The \textbf{second} row depicts the topology segmentation. The \textbf{third} row shows the disentangled wiring harness skeleton. The \textbf{fourth} row shows the disentangled segment lengths of the wiring harness. }
	\label{fig:wh_samples}
    %\hspace{-2mm}
    \vspace{-6mm}
\end{figure}

\subsection{PointWire Dataset}
The overview of the PointWire dataset generation pipeline is given in Fig.~\ref{fig:wh_pipeline}. This dataset is based on 40 real wiring harnesses used in the automotive sector. We use the high-precision industrial-grade Photoneo MotionCam-3D M, which is based on the previous PhoXi 3D Scanner M. A thorough study of its characteristics is presented in \cite{cop2021new}. To ensure a constant distance between the perception sensor and the wiring harness, we mount the 3D scanner on a robot arm and set the distance to the sensor's optimal distance based on \cite{cop2021new}.

First, a dense 3D scan of the wiring harness is obtained with the Photoneo MotionCam-3D M. An entangled wiring harness's captured point cloud scan contains more than 3 million sub-millimeter precise points. Secondly, the raw input point cloud is processed by removing the plane as well as the noise, using RANSAC and statistical outlier removal provided by Open3D \cite{zhou2018open3d}. The result is a clean point cloud of the wiring harness. The next step is the generation of a corresponding skeleton of the cleaned point cloud. We utilize the Laplacian-based contraction method \cite{laplace} to generate the baseline skeletons, and the result is shown in Fig.~\ref{fig:wh_pipeline}. Further, we manually purged and relocated the skeleton points to unambiguously define the bifurcation and end nodes.
The obtained graph is used for the rigging procedure in the 3D software suite Blender \cite{blender}, and for assigning the ground truth labels of the topological structure elements. Nodes with a degree greater than 2 are considered to be bifurcation nodes, while every node with a degree of 1 is an endpoint. The point annotation is then based on the thickness of the neighboring segments. Additionally, the segment lengths are calculated using a B-spline interpolation of the individual graph segments. Finally, by rigging and animating the cleaned point cloud, we extend each wiring harness point cloud scan to 300 samples. Each rigged real scan is manually moved in a physically feasible configuration. The in-between frames were interpolated by Blender as depicted in Fig.~\ref{fig:wh_pipeline}. Additional samples of entangled wiring harnesses are shown in Fig.~\ref{fig:wh_samples}. Furthermore, it demonstrates the challenging perception task setting that DLOs represent due to potentially complex entanglement.

\begin{figure}[t!]
    \centering
    \includegraphics[width=0.85\columnwidth]{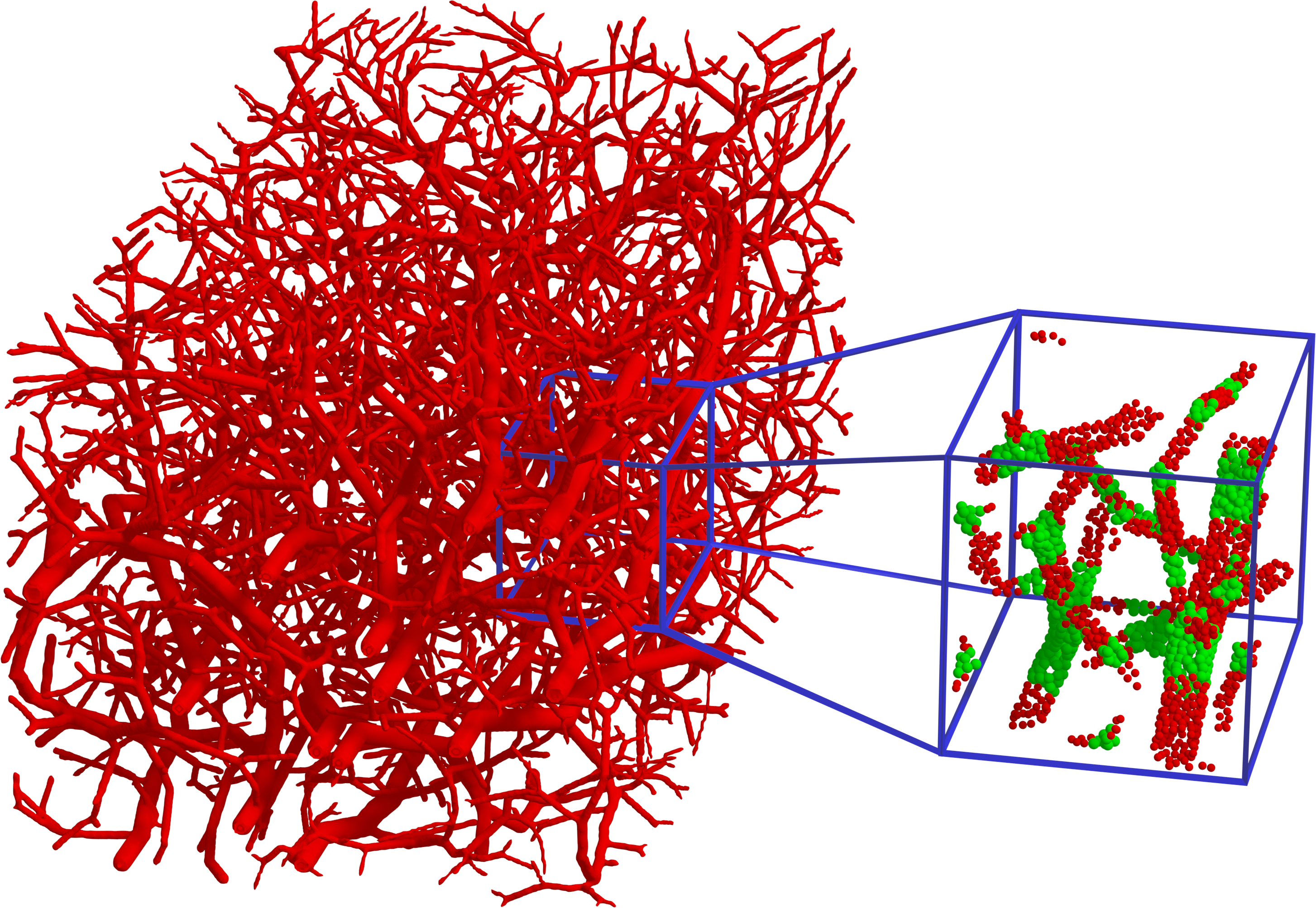}
    \caption{Blood vessel point cloud data generation. The entire synthetic blood vessel mesh is cropped into smaller subsets and uniformly sampled to 2048 points. Red points indicate the vessels, while the green points indicate the bifurcation. }
    \label{fig:dvn_mesh2pcl}
\end{figure}

\subsection{PointVessel Dataset}

The presented point cloud dataset of blood vessels, PointVessel, is derived from synthetic medical blood vessel data introduced in \cite{dvn}.

The original DeepVesselNet \cite{dvn} dataset is available as NIfTI files, which is a common data type in the medical imaging field. 
% \begin{figure*}[ht!]
%     \centering
%     \includesvg[width=0.59\textwidth]{./figures/pw_boxplot.svg} \hspace{-22mm}
%     \includesvg[width=0.52\textwidth]{./figures/pv_boxplot.svg} 
%     \caption{Statistics of the class distributions of the PointWire and PointVessel datasets. The PointWire dataset has a diverse distribution across the train, validation, and test sets. The PointVessel dataset has a relatively higher average of class instances per sample and an equal distribution across the sets.}
%     \label{fig:boxplot}
% \end{figure*}

\begin{figure*}
    \centering
    \includegraphics[width=0.95\textwidth]{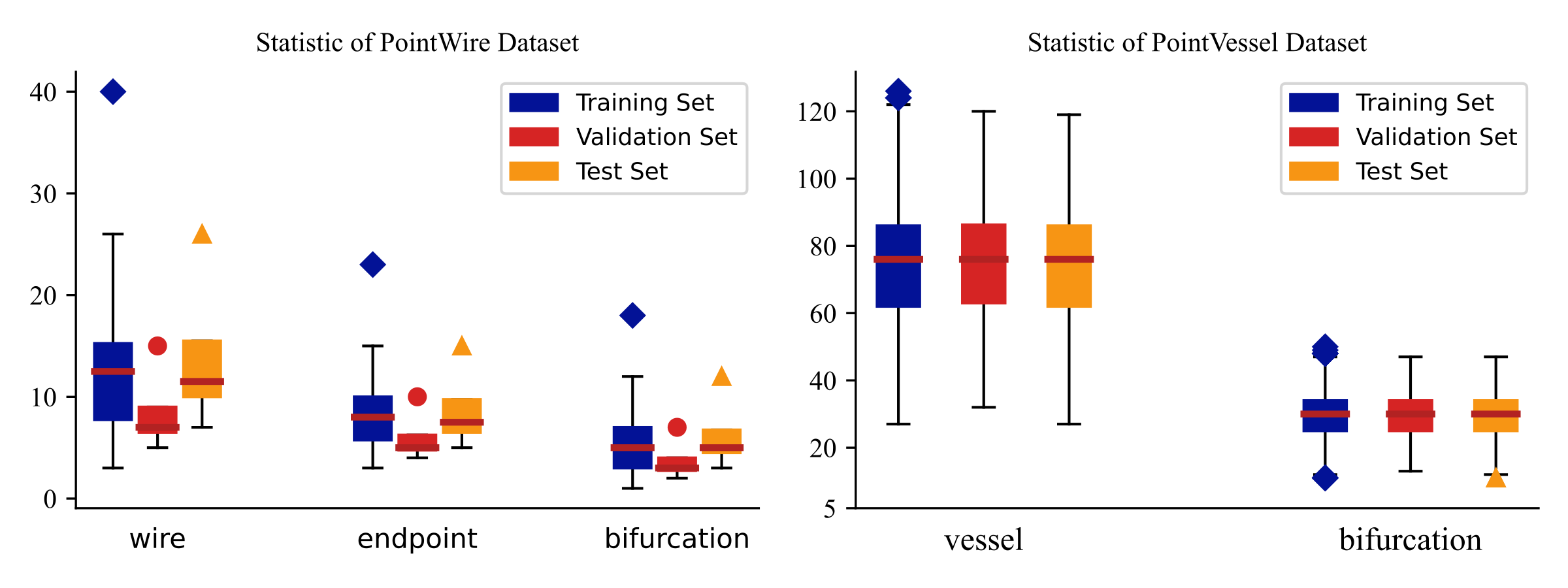}
    \vspace{-5mm}
    \caption{Statistics of the class distributions of the PointWire and PointVessel datasets. The PointWire dataset has a diverse distribution across the train, validation, and test sets. The PointVessel dataset has a relatively higher average of class instances per sample and an equal distribution across the sets.}
    \label{fig:boxplot}
\end{figure*}

There are 136 different blood vessel volumes with the size of $325 \times 304 \times 600$ voxels. The dataset processing pipeline is shown in Fig.~\ref{fig:dvn_mesh2pcl}. The first step is to convert the NIfTI blood vessel volumes using the vtk toolkit \cite{VTK4} into a 3D mesh. As each blood vessel volume contains a large number of vessels and bifurcations, we divide each volume into 96 same-sized overlaying crops, similarly to \cite{li2022vbnet}. The cropping strategy is depicted in Fig.~\ref{fig:dvn_mesh2pcl} with the blue box inside the blood vessel volume. Afterward, the surface mesh of each cropped volume is uniformly subsampled to a point cloud of 2048 points, shown in Fig.~\ref{fig:dvn_mesh2pcl} with the enlargened blue box. The ground truth bifurcation points and the vessel radii from \cite{dvn} are used to assign the corresponding point annotations. More details and figures regarding the PointVessel dataset are given in the supplementary.

\subsection{Statistics}
The PointWire dataset has 12000 samples, while the PointVessel dataset has 13056 samples. For both PointWire and PointVessel, each sample has 2048 points, and the train/val/test split is set to $80\% / 10\% / 10\%$. The PointWire dataset consists of the wire, bifurcation, and endpoint class. The PointVessel dataset has two classes: vessel and bifurcation. On average, PointWire and PointVessel have 26 and 104 instances per sample, respectively. Finally, the class distribution for both datasets is given in Fig.~\ref{fig:boxplot}. More details regarding the dataset statistics are given in the supplementary.

\section{Experiments}

This section provides the baseline benchmark experiments for our proposed two datasets, PointWire and PointVessel, on the topology segmentation task. Additionally, we conduct transferability experiments of the select state-of-the-art point cloud segmentation methods on the PointWire and PointVessel datasets. The intention is to analyze whether the methods are capable of learning general topological concepts of two different DLO object types, i.e., wiring harnesses and blood vascular systems.  

\subsection{Experiment Metrics}
\textbf{Point cloud segmentation evaluation metrics}
We evaluate different methods on our proposed datasets in terms of mean Intersection over Union (mIoU, \%), mean accuracy (mACC, \%), overall accuracy (OA, \%), Length-Mean Average Error (L-MAE), Segment-Mean Average Error (S-MAE), and IoU for each class: wire (w IoU), and bifurcation (b IoU). We calculate the L-MAE   a

\textbf{Disentanglement evaluation metrics}
We evaluate the disentanglement by using the mean absolute error to calculate the total length error (L-MAE) and the segment-wise error (S-MAE). L-MAE aggregates the lengths of all segments of a skeleton.
For S-MAE we do a bipartite matching of the bifurcation and endpoints. We then compare the individual segment lengths of the ground truth to the length of the path between the matched graph nodes.

\subsection{Implementation Details}
We evaluate PointNet++\cite{qi2017pointnet++}, DGCNN\cite{wang2019dynamic}, PCT\cite{guo2021pct}, CurveNet\cite{xiang2021walk}, DeltaConv\cite{Wiersma2022DeltaConv}, and RepSurf\cite{ran2022surface} on our large-scale benchmarks to get the baseline results for topological segmentation. Furthermore, we show the transferability performance of each network over the two datasets. For the baseline method of the disentanglement benchmark, we generate the nodes with semantic laplacian-based contraction \cite{meyer2023cherrypicker} and connect them with minimum spanning tree. All the experiments are conducted using one NVIDIA GeForce RTX 3090 (24Gb). For all network training on the two datasets, we set the batch size as 8 and modified the learning rate according to the Linear Scaling Rule \cite{goyal2017accurate}. Please refer to the supplementary material for more experimental setting details.
\begin{table}[ht!]
  \centering
  \setlength{\tabcolsep}{2pt}
  \resizebox{\columnwidth}{!}{ 
    \begin{tabular}{cccccccc}
    \toprule[1pt]
    \multicolumn{1}{c}{\textbf{Method}} & \textbf{mIoU}$\uparrow$ & \textbf{mAcc}$\uparrow$ & \textbf{OA}$\uparrow$ & \textbf{w IoU}$\uparrow$ & \textbf{e IoU}$\uparrow$ & \textbf{b IoU}$\uparrow$\\ \bottomrule[1pt]
        \multicolumn{1}{r}{PointNet++ \cite{qi2017pointnet++}} & 49.08 & 69.53 & 73.09 & 68.49 & 54.86 & 23.89 \\
    \multicolumn{1}{r}{DGCNN \cite{wang2019dynamic}} & 54.84 & 80.56 & 73.86 & 67.39 & 66.90 & 30.25 \\
    \multicolumn{1}{r}{PCT \cite{guo2021pct}} & 36.68 & 59.27 & 62.01 & 56.59 & 30.64 & 22.8  \\
    \multicolumn{1}{r}{CurveNet \cite{xiang2021walk}} & 53.92 & 68.96 & 79.91 & 76.92 & 53.31 & 31.53 \\
    \multicolumn{1}{r}{DeltaConv \cite{Wiersma2022DeltaConv}} & 60.47 & 81.06 & 80.34 & 75.64 & 70.49 & 35.29 \\
    \multicolumn{1}{r}{RepSurf \cite{ran2022surface}} & \textbf{73.6} & \textbf{85.6} & \textbf{85.22} & \textbf{81.52} & \textbf{72.72} & \textbf{40.14} \\ \bottomrule[1pt]
    \end{tabular}
}
  \caption{PointWire (all classes) benchmark. We evaluate different methods on the test set.}
  \label{tab:wh_all}
\end{table}

\subsection{PointWire Benchmark}

\textbf{PointWire (all classes)} The quantitative evaluation of the chosen point cloud segmentation methods on the PointWire all classes are given in Tab.~\ref{tab:wh_all}. RepSurf \cite{ran2022surface} and DeltaConv \cite{Wiersma2022DeltaConv} outperform them by a large margin the other methods. The reason for that lies in the point cloud surface learning strategies used by RepSurf and DeltaConv. By utilizing local 2D manifold feature aggregation, these methods are able to generalize well on the challenging PointWire dataset. Comparing the qualitative results of the evaluation on the PointWire test set, given in the first row of Fig.~\ref{fig:base_quali}, we can observe that both RepSurf and DeltaConv correctly segment the bifurcations and endpoints with the least amount of false positives. PointNet++ \cite{qi2017pointnet++} and DGCNN \cite{wang2019dynamic} heavily under-segmented the bifurcations, while PCT \cite{guo2021pct} and CurveNet \cite{xiang2021walk} have relatively more false positives located at wire segments. However, based on the results from Tab.~\ref{tab:wh_all}, it is clear that the PointWire dataset is rather challenging, especially for the bifurcation class since the highest IoU score is $40.14~\%$, achieved by RepSurf \cite{ran2022surface}.

\textbf{PointWire (bifurcation)} We show the quantitative results of PointWire bifurcation in Tab.~\ref{tab:tab3}. The reason for conducting the PointWire$_{bifurcation}$ experiment is to have an identical setting, w.r.t.~the same number of classes, as in the PointVessel benchmark. Similar to the results on PointWire all classes benchmark, the surface-based method DeltaConv performs best on the PointWire bifurcation experiment. Compared with the PointWire$_{all}$, CurveNet and PointNet++ achieve better performance on the PointWire$_{bifurcation}$ benchmark in terms of mIoU metrics. Moreover, by comparing the qualitative results shown in the second row in Fig.~\ref{fig:base_quali} DeltaConv and PointNet++ have the visually best results. CurveNet fails to identify bifurcation segments and collapses into a mode of segmenting almost the entire wiring harness as the wire class. The other methods \cite{wang2019dynamic, guo2021pct, ran2022surface} have the tendency to under-segment the bifurcation class, which results in less overall accuracy (OA).

\subsection{PointVessel Benchmark}
The quantitative and qualitative results on the PointVessel benchmark of the state-of-the-art point cloud segmentation methods are reported in Tab.~\ref{tab:tab3}. Overall, the selected methods achieve better mAcc and mIoU compared to the PointWire benchmark. This indicates that the PointWire benchmark presents a more challenging topology segmentation task. The reasons for that are twofold: 1) the set distributions are similar, and 2) the PointVessel dataset is synthetic with no noise and a complete surface of the underlying structures. Surprisingly, PointNet++ \cite{qi2017pointnet++} achieves together with CurveNet \cite{xiang2021walk} the best results, as reported in Tab.~\ref{tab:tab3}. This confirms the finding of a recent study \cite{qian2022pointnext} that shows that with the right settings, PointNet++ can still match state-of-the-art methods. As mentioned above, the reason for such a good performance of CurveNet and PointNet++ is very likely due to the way how the feature aggregation is performed, and which is more suitable for clean, i.e.~synthetic data. Especially, the curve feature aggregation approach introduced by CurveNet \cite{xiang2021walk} achieves the best results on smooth and continuous point cloud surfaces.

If we take a look at the bifurcation class quantitative and qualitative results in Tab.~\ref{tab:tab3} and the first row of Fig.~\ref{fig:transfer_quali}, it is obvious that again the surface learning-based methods perform reasonably well. On the other hand, PCT \cite{guo2021pct} achieves the worst results. The main reason for this is mainly due to the fact that transformers have the huge disadvantage of requiring massive amounts of data. One interesting observation is that all the models have issues in correctly segmenting bifurcations on larger blood vessels. However, bifurcations located at blood vessels with a smaller diameter are successfully segmented by all the state-of-the-art methods.

\subsection{Transferability Benchmark}
%%%%%% WH all classes %%%%%%
%\input{ieeeconf/segmentation_res}
%\input{ieeeconf/new_experiment_results_table}
%\input{ieeeconf/qualitative_results}
\begin{table*}[h!]
  \centering
  \resizebox{\textwidth}{!}{
    \begin{tabular}{ccccccccccccc}
        \toprule[1pt]
        \multirow{3}{*}{\textbf{Benchmark}} & 
        \multicolumn{5}{c}{PointWire$_{bifurcation}$} & 
        \multicolumn{5}{c}{PointVessel} & \multicolumn{2}{c}{PointWire$_{bifurcation}$} \\ 
        & \multicolumn{5}{c}{PointVessel $\rightarrow$ PointWire$_{bifurcation}$} & \multicolumn{5}{c}{PointWire$_{all}$ $\rightarrow$ PointVessel}  & \multicolumn{2}{c}{PointVessel}  \\ 
        & \multicolumn{5}{c}{-} & \multicolumn{5}{c}{PointWire$_{bifurcation}$ $\rightarrow$ PointVessel}  & \multicolumn{2}{c}{-} \\ \bottomrule[1pt]
        \multicolumn{1}{c}{\textbf{Method}} & \textbf{mIoU}$\uparrow$ & \textbf{mAcc}$\uparrow$ & \textbf{OA}$\uparrow$ & \textbf{w IoU}$\uparrow$ & \textbf{b IoU}$\uparrow$ & \textbf{mIoU}$\uparrow$ & \textbf{mAcc}$\uparrow$ & \textbf{OA}$\uparrow$ & \textbf{v IoU}$\uparrow$ & \textbf{b IoU}$\uparrow$ & \textbf{L-MAE}$\downarrow$ & \textbf{S-MAE}$\downarrow$ \\ \bottomrule[1pt]
        % pointnet++
        \multirow{3}{*}{PointNet++ \cite{qi2017pointnet++} (2017)}  
        & 56.74 & 74.22 & 84.36 & 83.22 & 30.27 & 84.79 & 94.32 & 92.34 & 88.64 & 80.93 & 0.89 & 0.48 \\
        & \underline{47.98} & 75.20 & 71.71 & 68.94 & 27.02 & 31.27 & 42.48 & 52.21 & 52.81 & 9.73 & 0.98 & 0.27 \\
        & - & - & - & - & - & 43.83 & 57.91 & 70.64 & 68.43 & 19.23 & - & - \\ \bottomrule[1pt]
        % dgcnn
        \multirow{3}{*}{DGCNN \cite{wang2019dynamic} (2019)}
        & 47.33 & 79.81 & 71.68 & 68.53 & 26.12 & 77.44 & 89.15 & 88.21 & 83.19 & 71.70 & \textbf{0.79} & 0.48 \\
        & 30.33 & 48.11 & 55.4 & 50.85 & 9.81 & 52.02$^{*}$ & 66.70 & 72.21 & 66.70 & 37.35 & 0.99 & 0.27\\
        & - & - & - & - & - & 45.94 & 66.08 & 63.67 & 52.36 & 39.51 & - & - \\ \bottomrule[1pt] 
        % pct
        \multirow{3}{*}{PCT \cite{guo2021pct} (2021)}
        & 46.64 & 65.55 & 75.51 & 73.99 & 19.29 & 57.38 & 76.70 & 73.62 & 63.46 & 51.31 & 0.90 & 0.49 \\
        & 35.31 & 51.31 & 65.71 & 60.19 & 10.42 & 40.34 & 55.02 & 62.9 & 58.51 & 22.17 & 1.01 & \textbf{0.26} \\
        & - & - & - & - & - & 38.14 & 54.95 & 57.36 & 49.46 & 26.82 & - & - \\ \bottomrule[1pt]
        % curvenet
        \multirow{3}{*}{CurveNet \cite{xiang2021walk} (2021)}
        & 57.3 & 67.86 &87.63 & 87.02 & 27.57 & \underline{84.99} & 93.02 & 92.63 & 89.38 & 80.60 & 0.87 & 0.49 \\
        & 45.5 & 53.54 & 81.86 & 81.51 & 9.49 & 41.41 & 55.81 & 69.52 & 67.78 & 15.05 & 0.99 & 0.27 \\
        & - & - & - & - & - & 40.49 & 55.13 & 69.62 & 68.21 & 12.77 & - & - \\ \bottomrule[1pt]
        % deltaconv
        \multirow{3}{*}{DeltaConv \cite{Wiersma2022DeltaConv} (2022)}
        & \textbf{59.22} & 81.09 & 84.47 & 83.03 & 35.41 & 84.64 & 92.75 & 92.45 & 89.15 & 80.13 & 0.86 & 0.46\\
        & 37.72 & 63.96 & 62.17 & 59.17 & 16.27 & 46.03 & 59.88 & 71.16 & 68.31 & 23.75 & 0.98 & 0.27 \\
        & - & - & - & - & - & \textbf{46.14} & 59.97 & 71.4 & 68.62 & 23.66 & - & - \\ \bottomrule[1pt]
        % repsurf
        \multirow{3}{*}{RepSurf \cite{ran2022surface} (2022)} 
        & 52.94 & 83.81 & 77.97 & 74.78 & 31.10 & 79.1 & 90.24 & 89.19 & 84.48 & 73.72 & 0.87 & 0.49\\ 
        & 37.04 & 50.72 & 79.89 & 71.21 & 2.87 & 44.27 & 58.30 & 69.68 & 66.91 & 21.63 & 0.99 & 0.27\\
        & - & - & - & - & - & 42.05 & 62.68 & 59.74 & 47.63 & 36.47 & - & - \\ \bottomrule[1pt]
    \end{tabular}
  }
  \caption{PointWire bifurcation benchmark, PointVessel benchmark, and transferability benchmark. \textbf{PointWire$_{bifurcation}$}: experiments using only the wire and bifurcation classes. \textbf{PointVessel $\rightarrow$ PointWire$_{bifurcation}$}: transferability evaluations on the PointWire$_{bifurcation}$ test set. \textbf{PointWire$_{all}$ $\rightarrow$ PointVessel} and \textbf{PointWire$_{bifurcation}$ $\rightarrow$ PointVessel}: transferability evaluations on the PointVessel test set. $ \_, *$ and \textbf{bold} denote the best results for each benchmark.}
  \label{tab:tab3}
\end{table*}

\begin{figure*}[]
	\centering 
    \setlength{\tabcolsep}{1pt}
    \renewcommand{\arraystretch}{2}
	\begin{tabular}{cccccccc} 
	%%%%%%%%%%%%%%%%%%%%%%%%%%%%%%%%%%%%%%%%%%%%%%%%%%%%%%%%%%%%%%%%%%%%%%%%%%%%%%%%%%%%%%%%%%%%%%%%%%%%%%%%%%%%%%%
	   ~ & \textit{Ground Truth} & \textit{PointNet++} & \textit{DGCNN} & \textit{PCT} & \textit{CurveNet} & \textit{DeltaConv} & \textit{RepSurf} \\
	%%%%%%%%%%%%%%%%%%%%%%%%%%%%%%%%%%%%%%%%%%%%%%%%%%%%%%%%%%%%%%%%%%%%%%%%%%%%%%%%%%%%%%%%%%%%%%%%%%%%%%%%%%%%%%%
		\rotatebox[y=12mm]{90}{\textit{PW$_{a}$}} &
		\includegraphics[width=.135\linewidth]{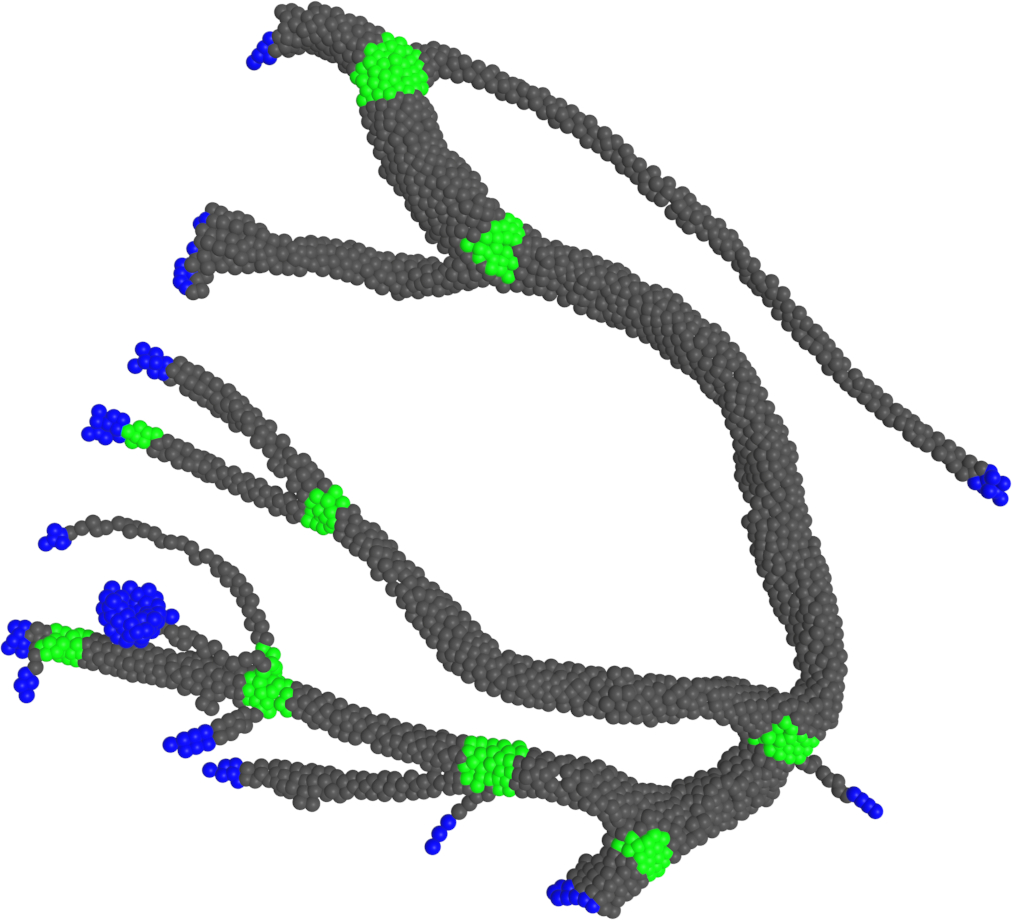} &
		\includegraphics[width=.135\linewidth]{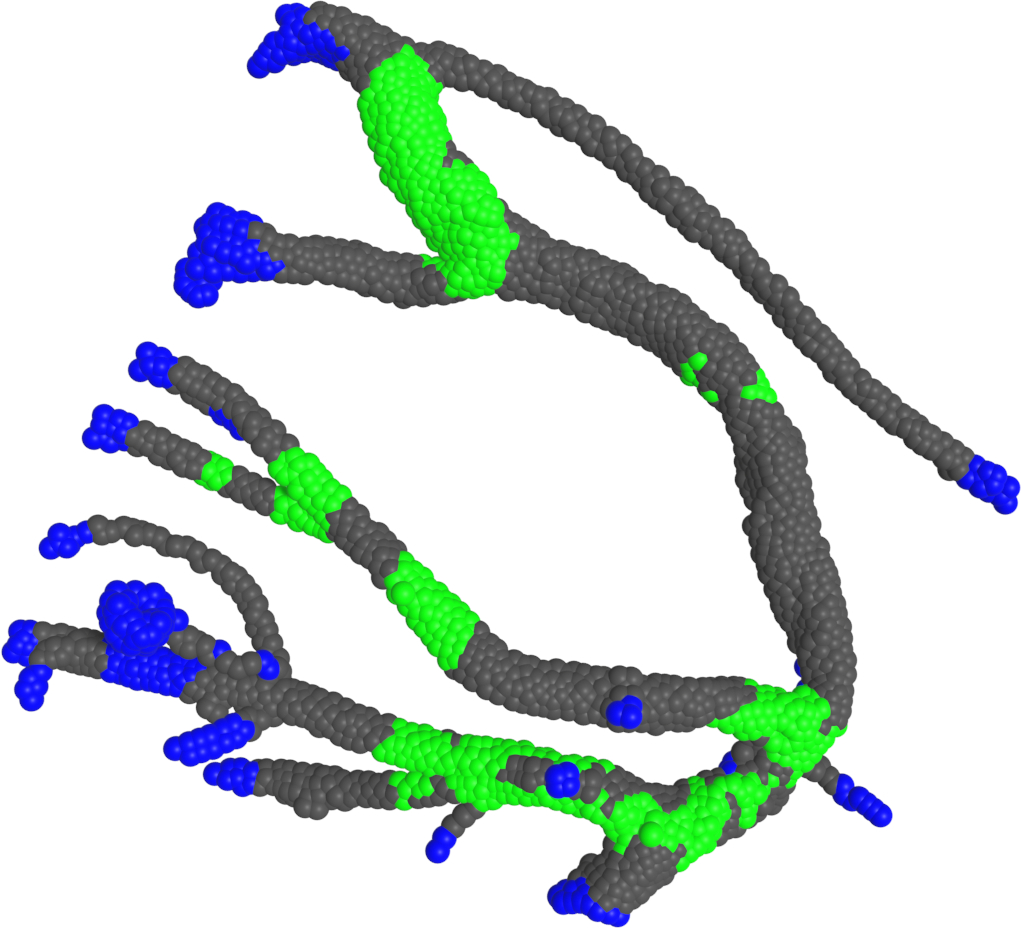} & 
		\includegraphics[width=.135\linewidth]{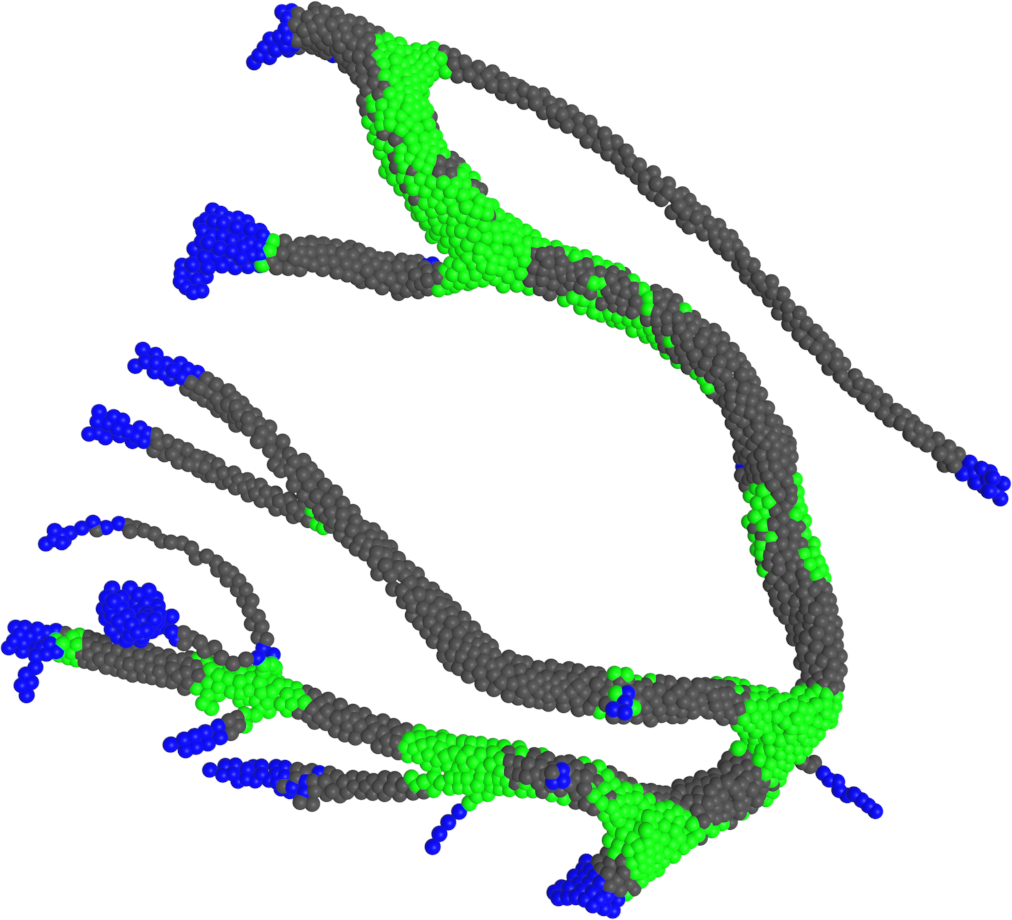} &  
		\includegraphics[width=.135\linewidth]{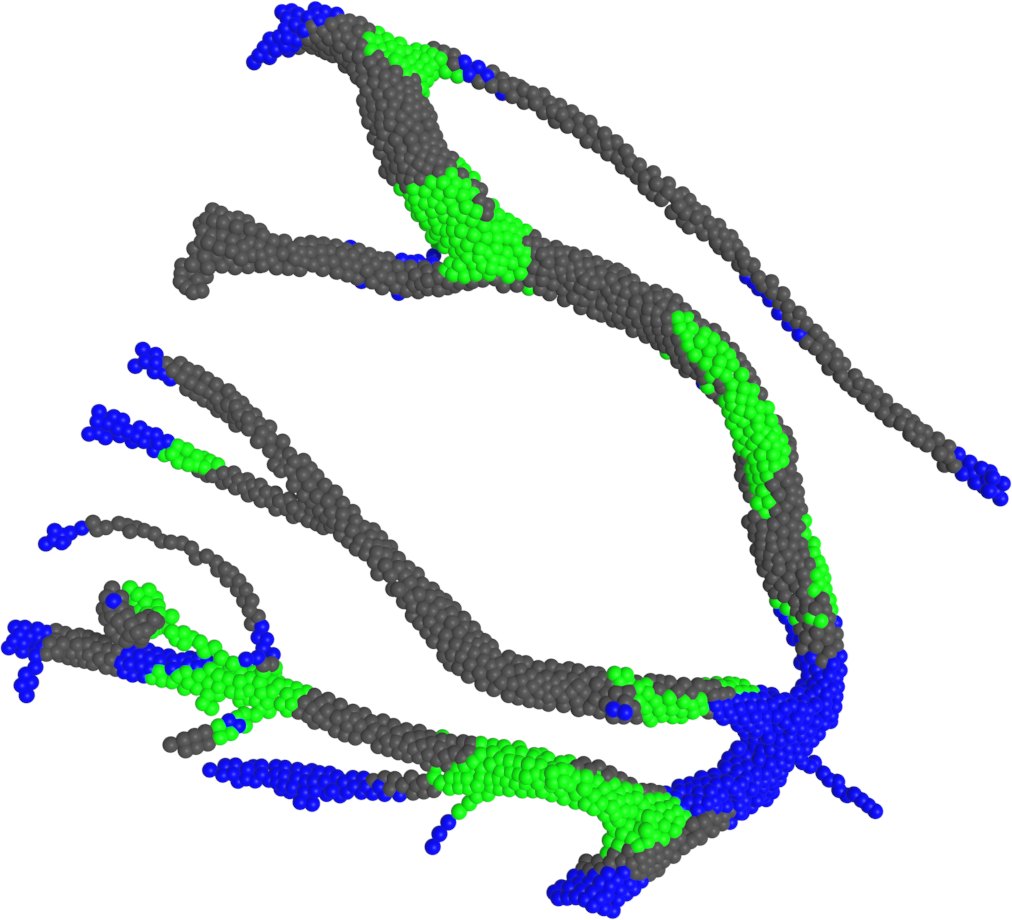} &  
		\includegraphics[width=.135\linewidth]{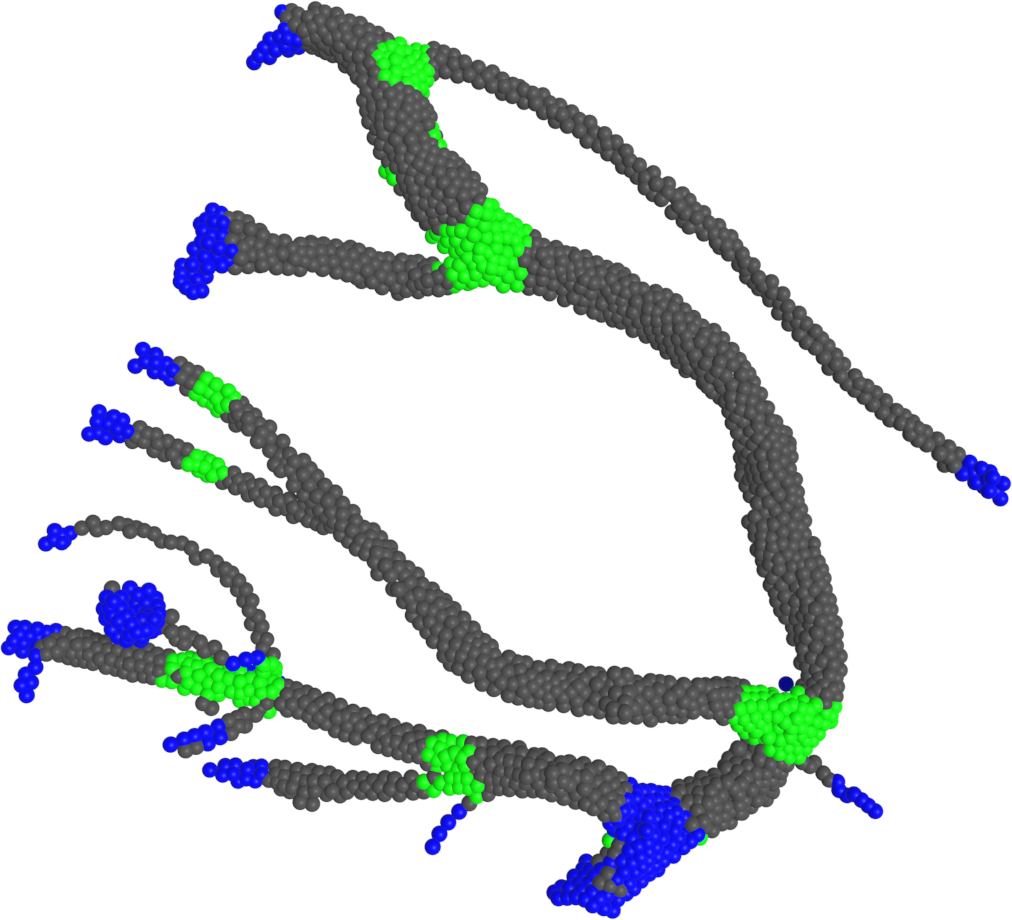} &  
		\includegraphics[width=.135\linewidth]{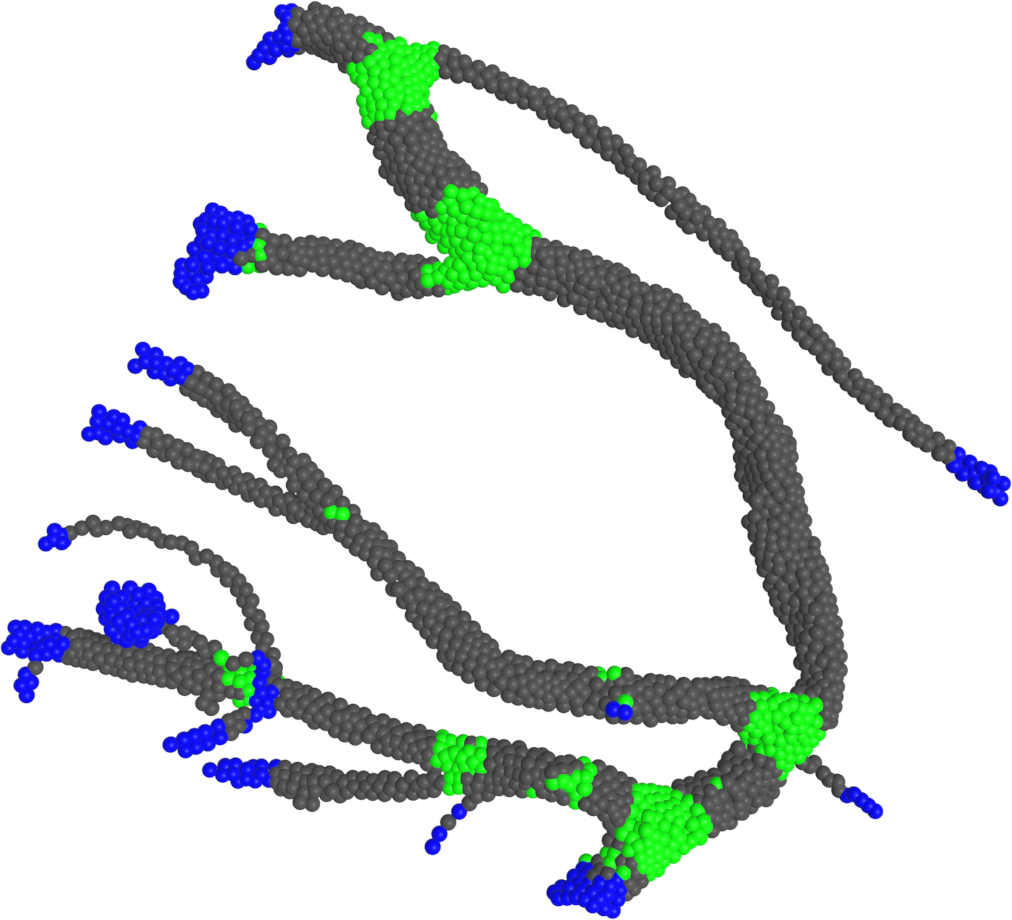} & 
		\includegraphics[width=.135\linewidth]{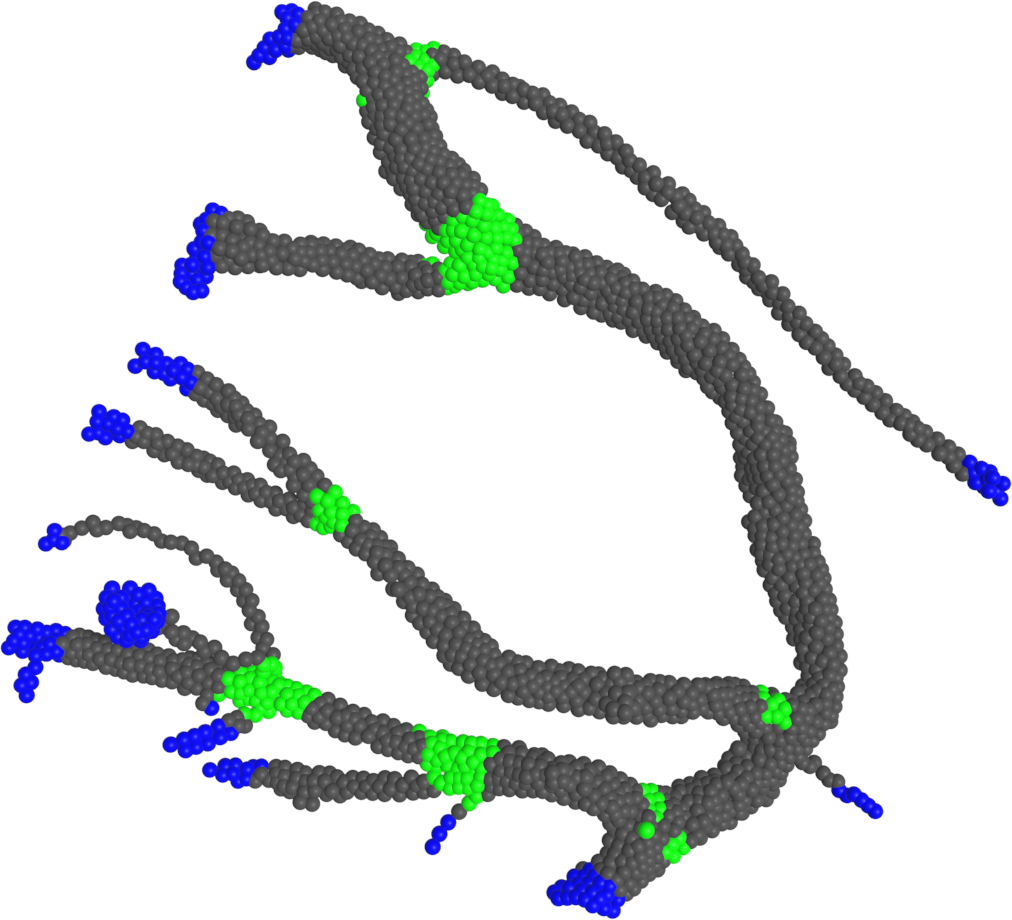} 
		%\vspace{-1mm} 
		\\ 
        \rotatebox[y=12mm]{90}{\textit{PW$_{b}$}} &
		\includegraphics[width=.135\linewidth]{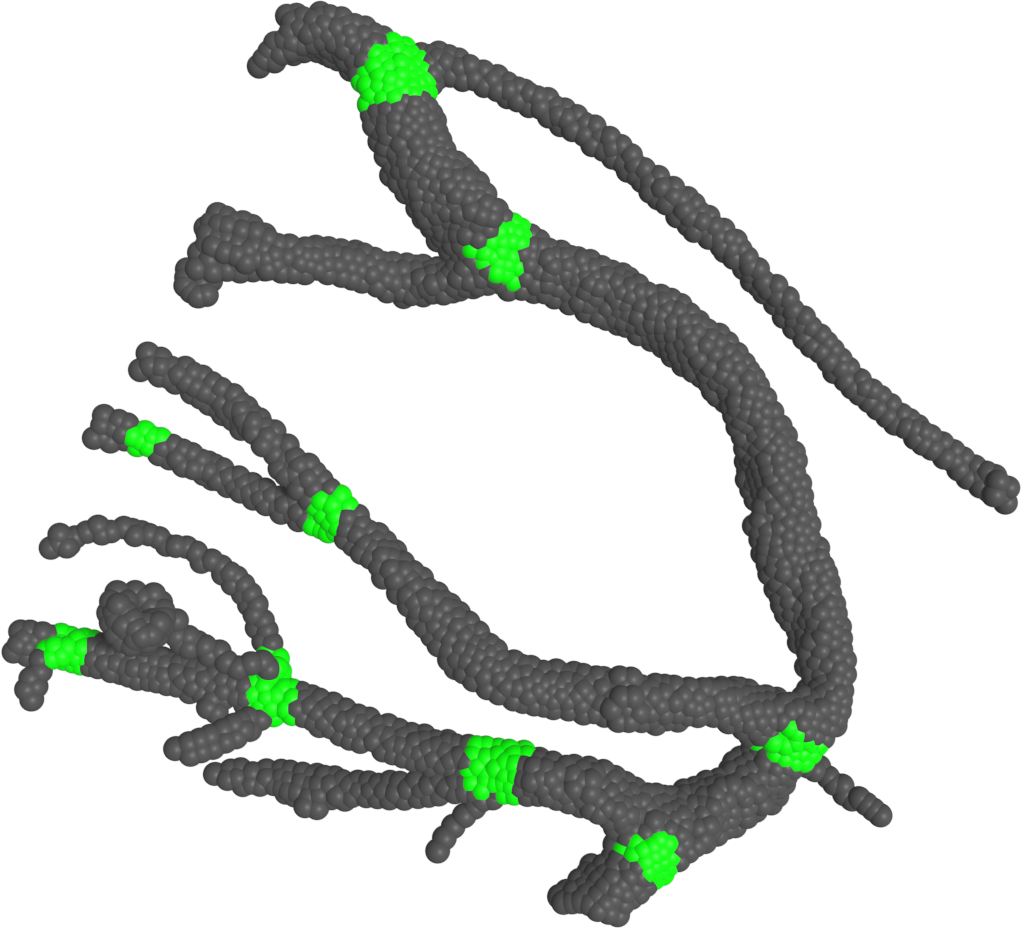} &
		\includegraphics[width=.135\linewidth]{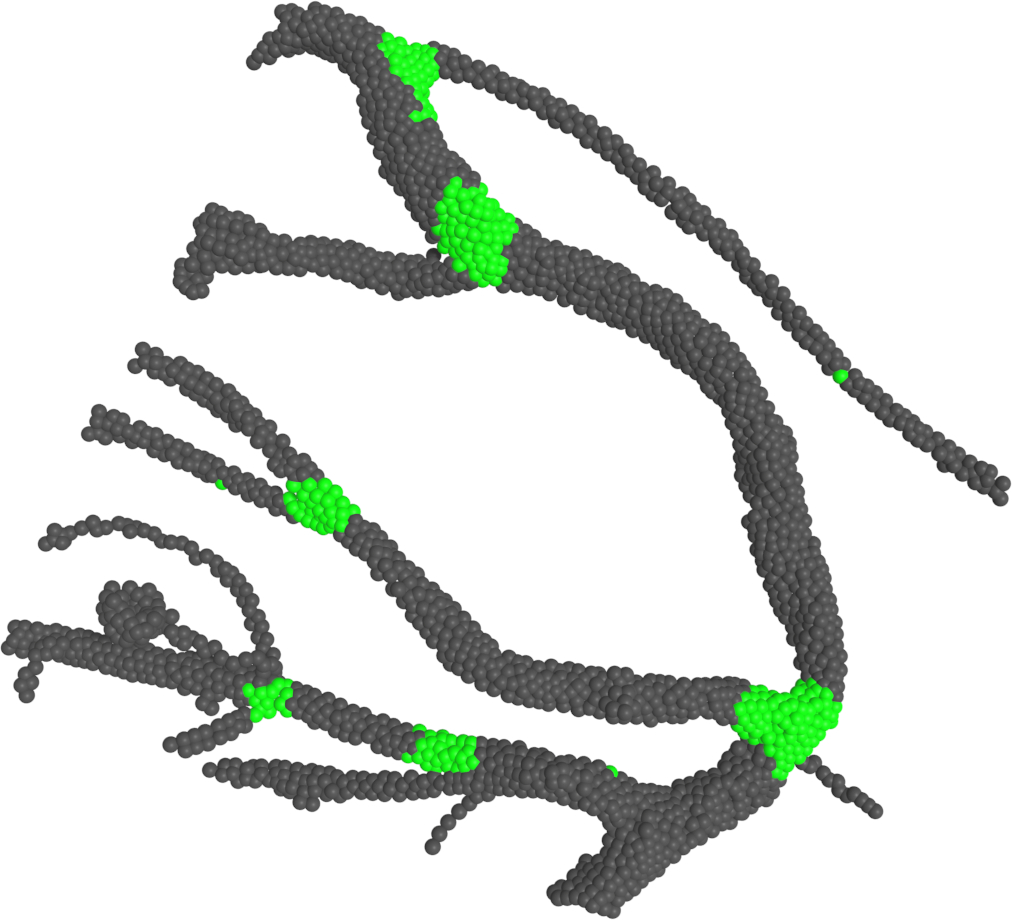} & 
		\includegraphics[width=.135\linewidth]{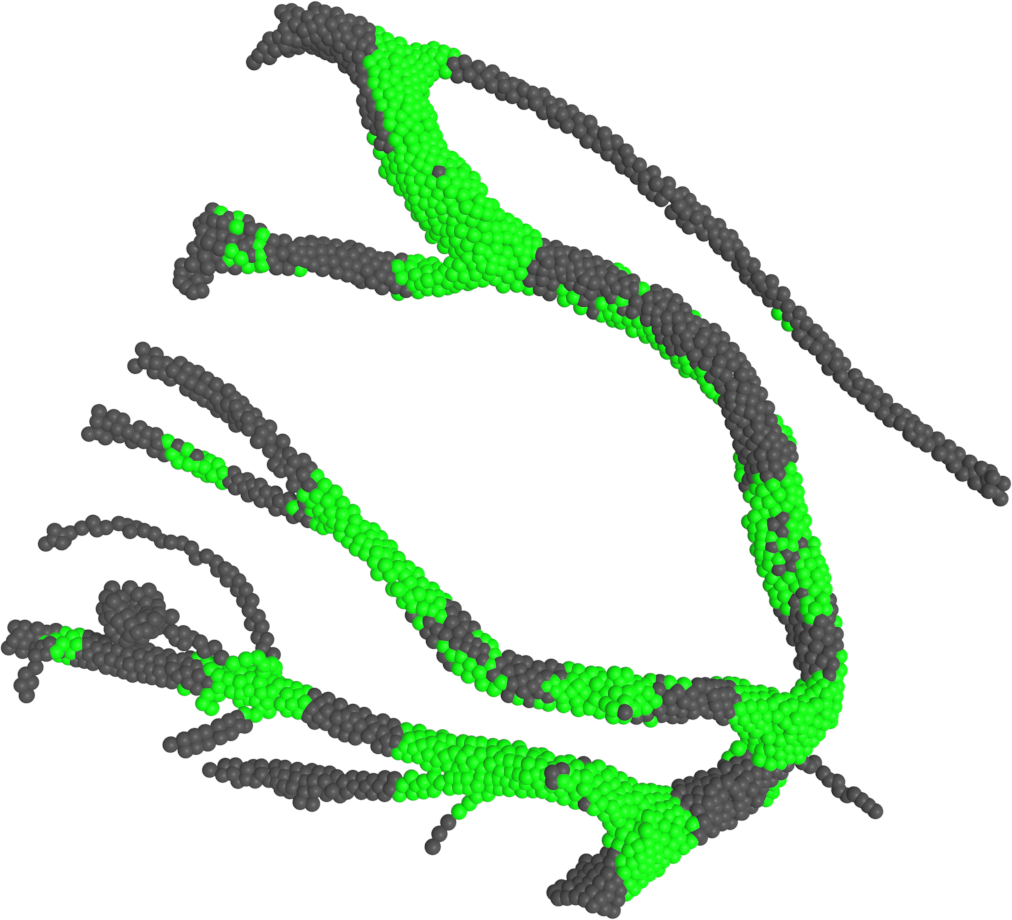} &  
		\includegraphics[width=.135\linewidth]{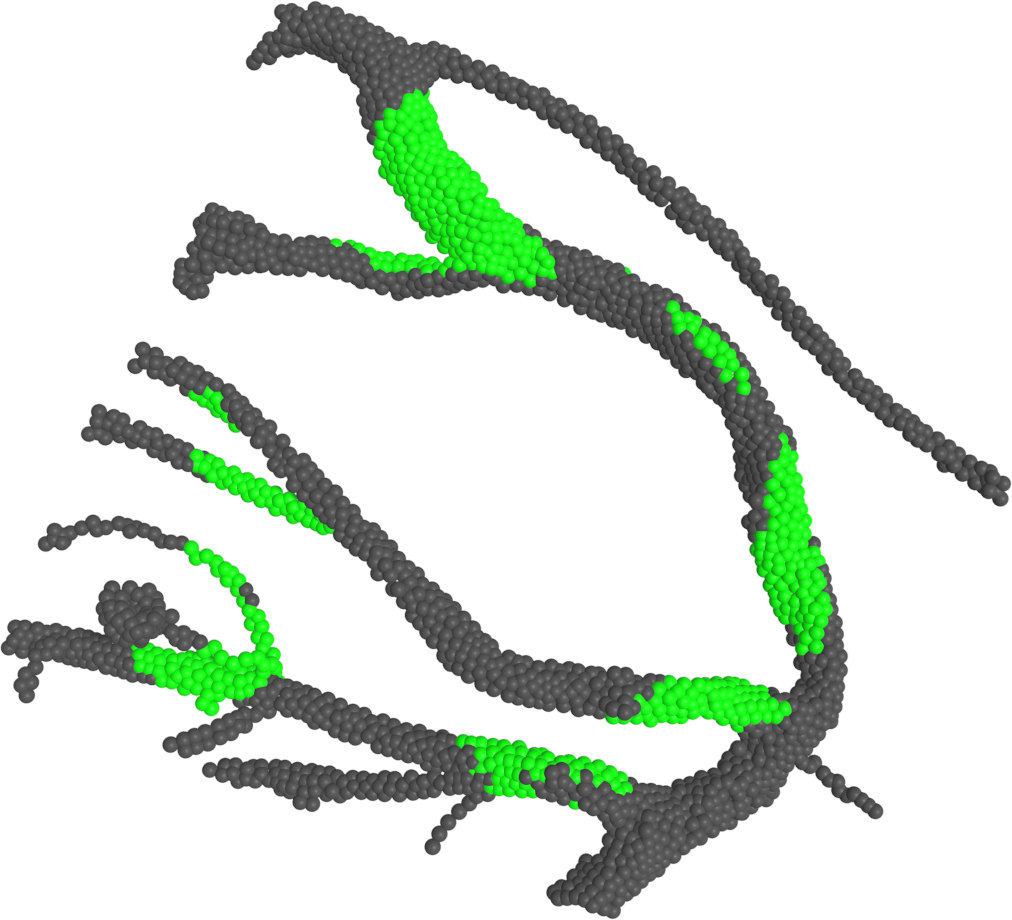} &  
		\includegraphics[width=.135\linewidth]{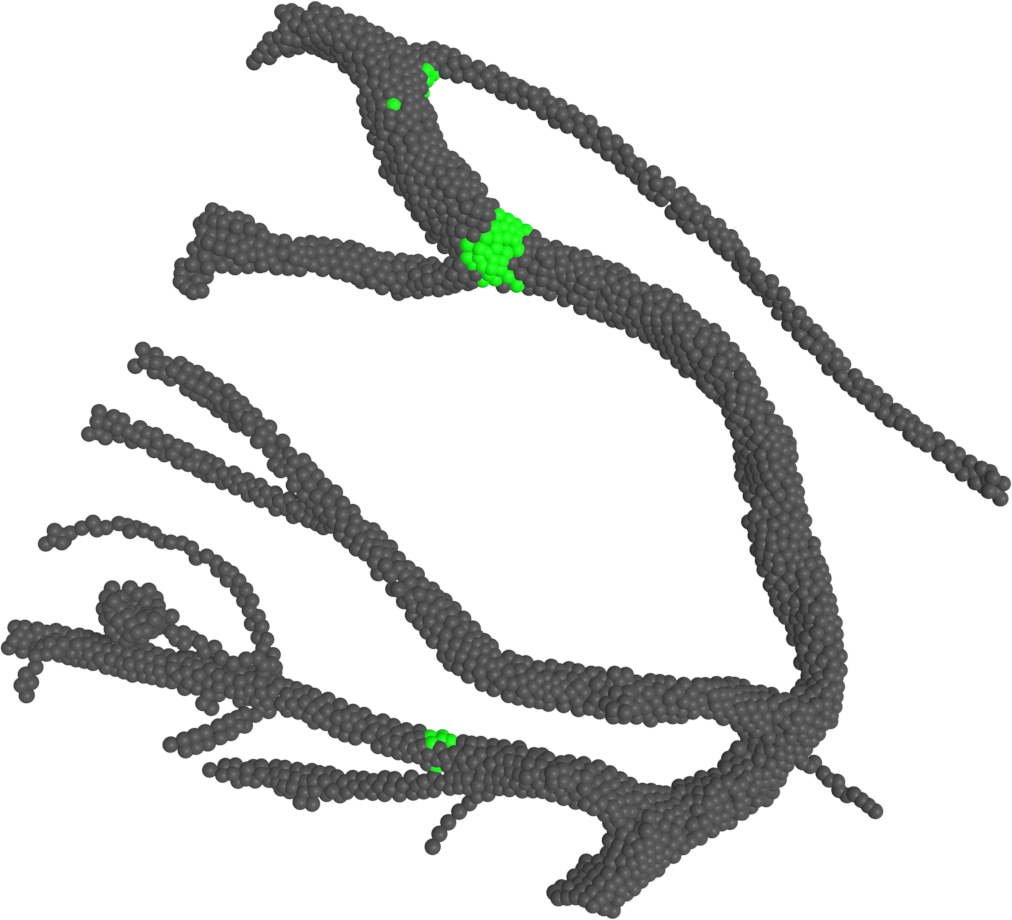} & 
		\includegraphics[width=.135\linewidth]{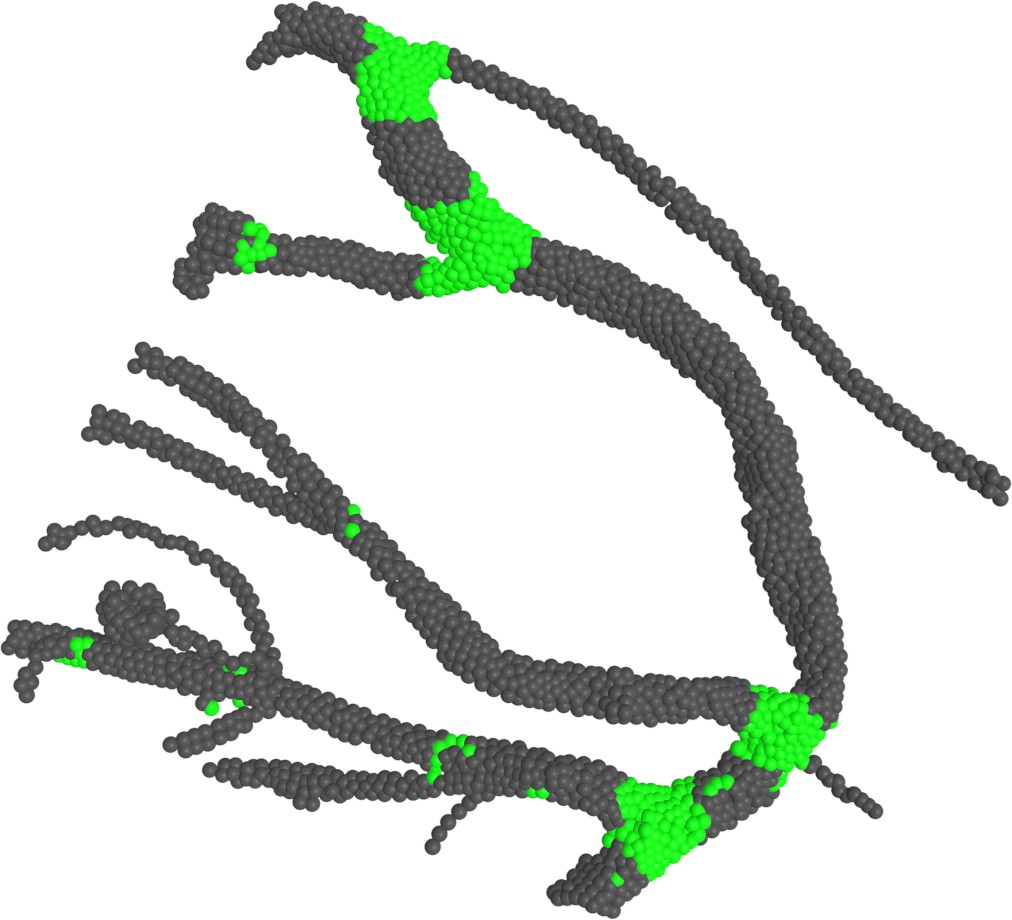} & 
		\includegraphics[width=.135\linewidth]{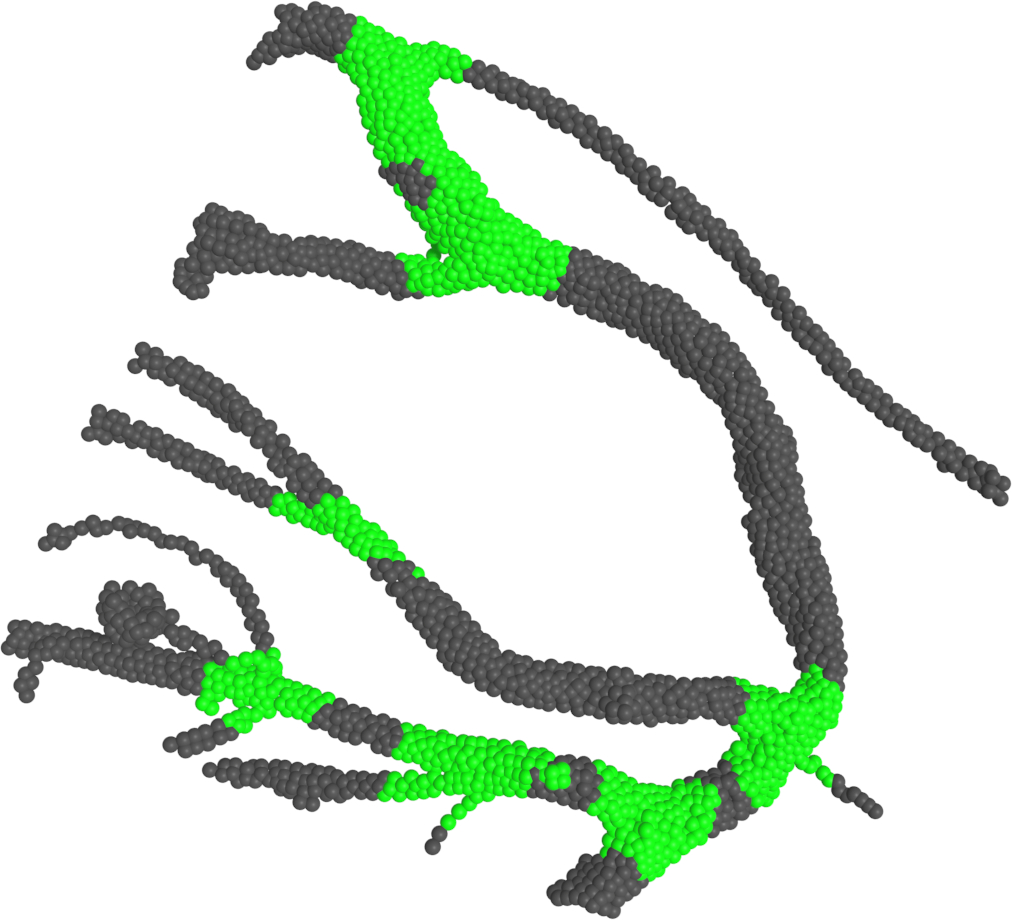}  
		%\vspace{-1mm} 
		\\ 
		\rotatebox[y=11mm]{90}{\textit{PV $\rightarrow$ PW$_{b}$}} &
		\includegraphics[width=.135\linewidth]{./figures/quali/gt_wh_bif.jpg} &
		\includegraphics[width=.135\linewidth]{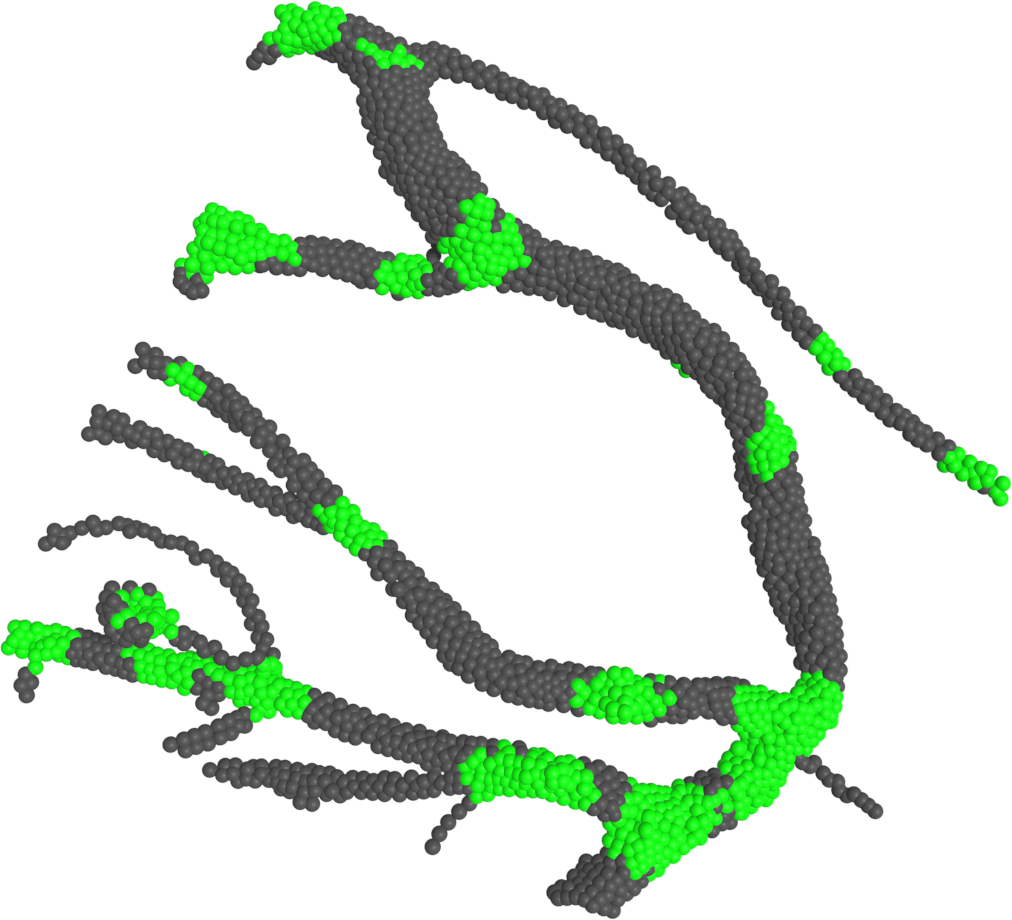} & 
		\includegraphics[width=.135\linewidth]{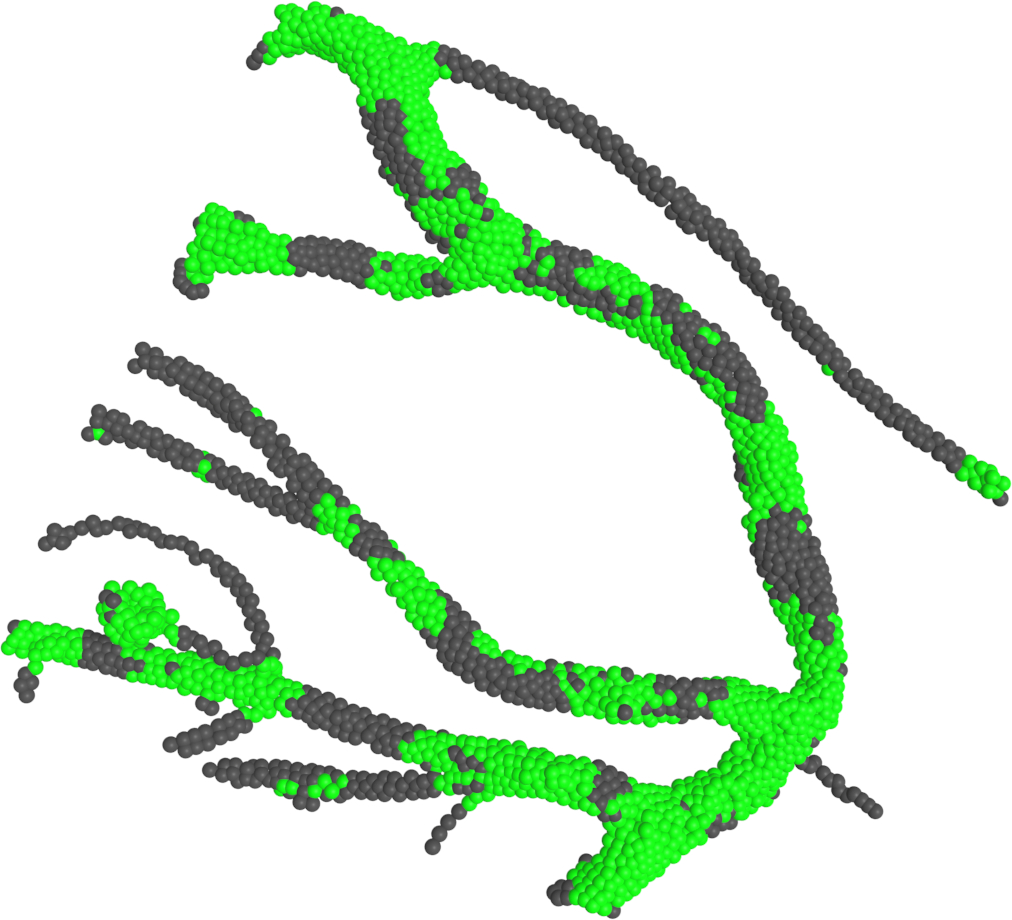} &  
		\includegraphics[width=.135\linewidth]{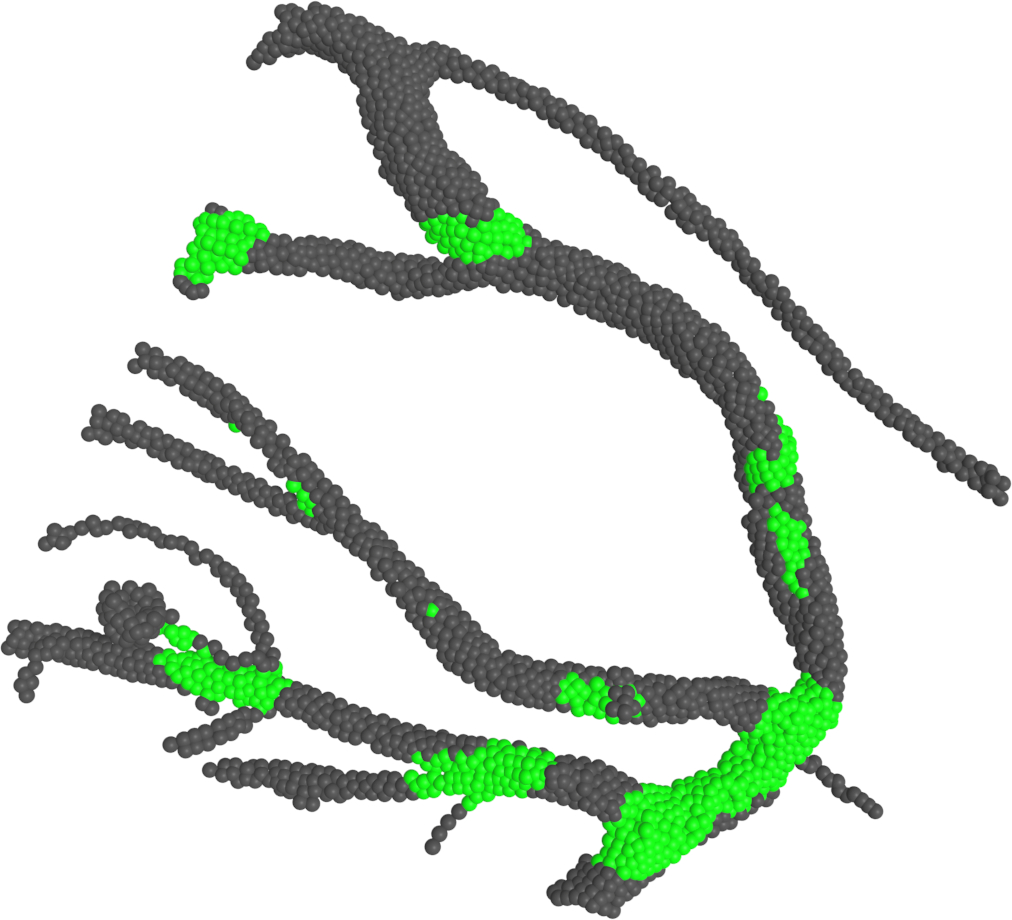} &  
		\includegraphics[width=.135\linewidth]{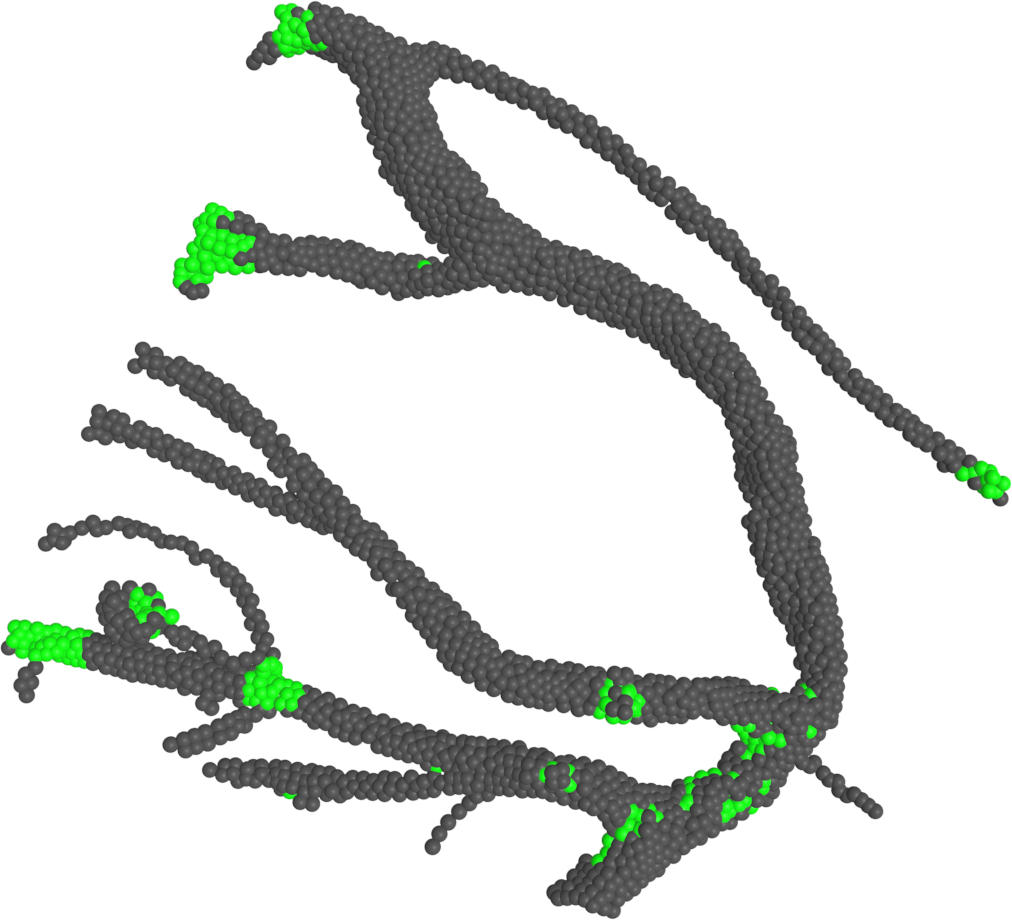} & 
		\includegraphics[width=.135\linewidth]{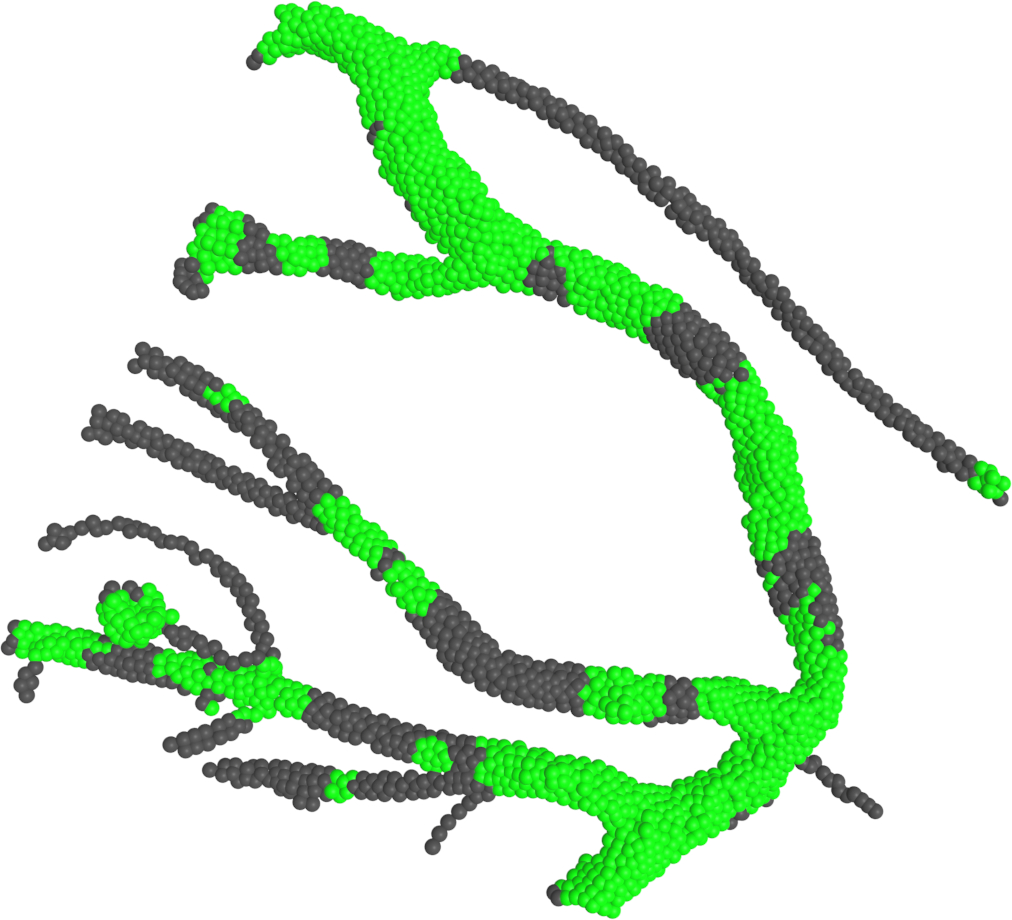} & 
		\includegraphics[width=.135\linewidth]{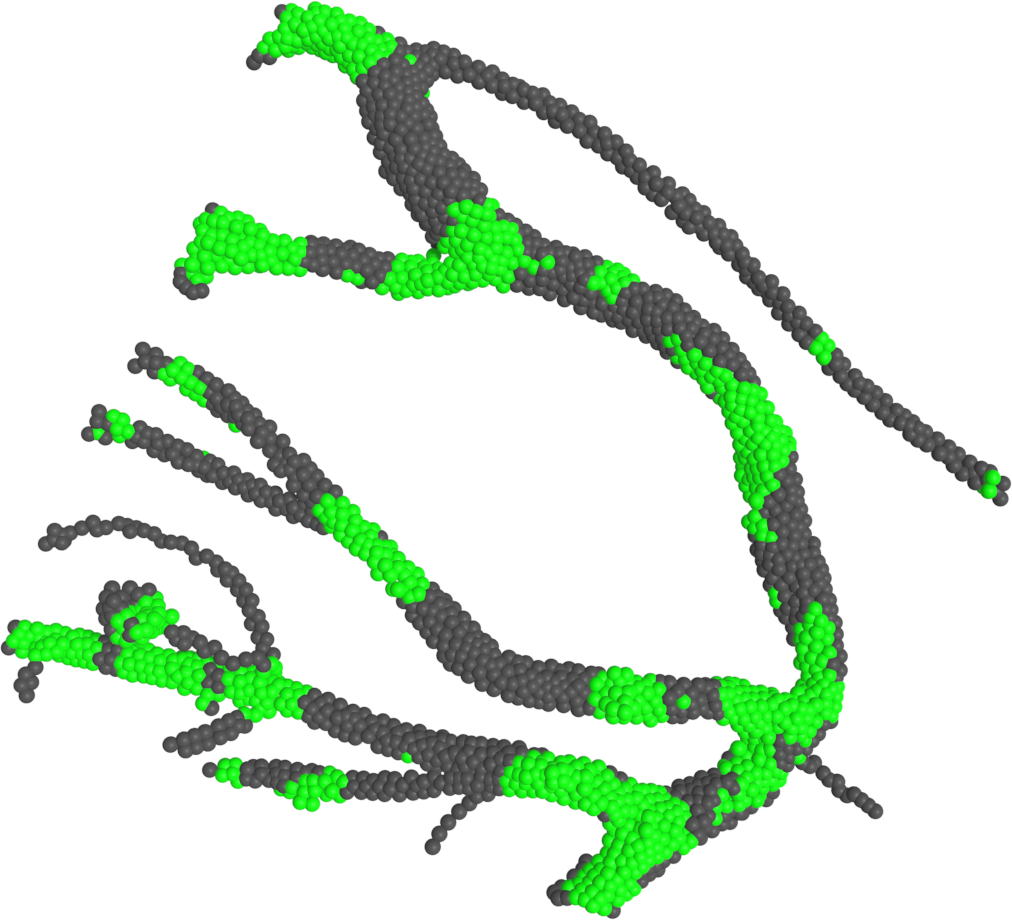} 
		%\vspace{-1mm}
    \end{tabular} 
	\caption{Qualitative results on PointWire all classes (PW$_{a}$), PointWire bifurcation (PW$_{b}$) and the transferability benchmark on the PointWire bifurcation (PV $\rightarrow$ PW$_{b}$). For PW$_{a}$ and PW$_{b}$,  \textcolor[RGB]{110, 110, 110}{gray} points are the wires, endpoints are \textcolor{blue}{blue}, and \textcolor{green}{green} ones are bifurcations.}
	\label{fig:base_quali}
    %\vspace{-5mm}
\end{figure*}

\begin{figure*}[]
	\centering 
	\renewcommand{\arraystretch}{2}
    \setlength{\tabcolsep}{1pt}
	\begin{tabular}{cccccccc} 
	%%%%%%%%%%%%%%%%%%%%%%%%%%%%%%%%%%%%%%%%%%%%%%%%%%%%%%%%%%%%%%%%%%%%%%%%%%%%%%%%%%%%%%%%%%%%%%%%%%%%%%%%%%%%%%%
	   ~ & \textit{Ground Truth} & \textit{PointNet++} & \textit{DGCNN} & \textit{PCT} & \textit{CurveNet} & \textit{DeltaConv} & \textit{RepSurf} \\
	%%%%%%%%%%%%%%%%%%%%%%%%%%%%%%%%%%%%%%%%%%%%%%%%%%%%%%%%%%%%%%%%%%%%%%%%%%%%%%%%%%%%%%%%%%%%%%%%%%%%%%%%%%%%%%%
		\rotatebox[y=11mm]{90}{\textit{PV}} &
		\includegraphics[width=.135\linewidth]{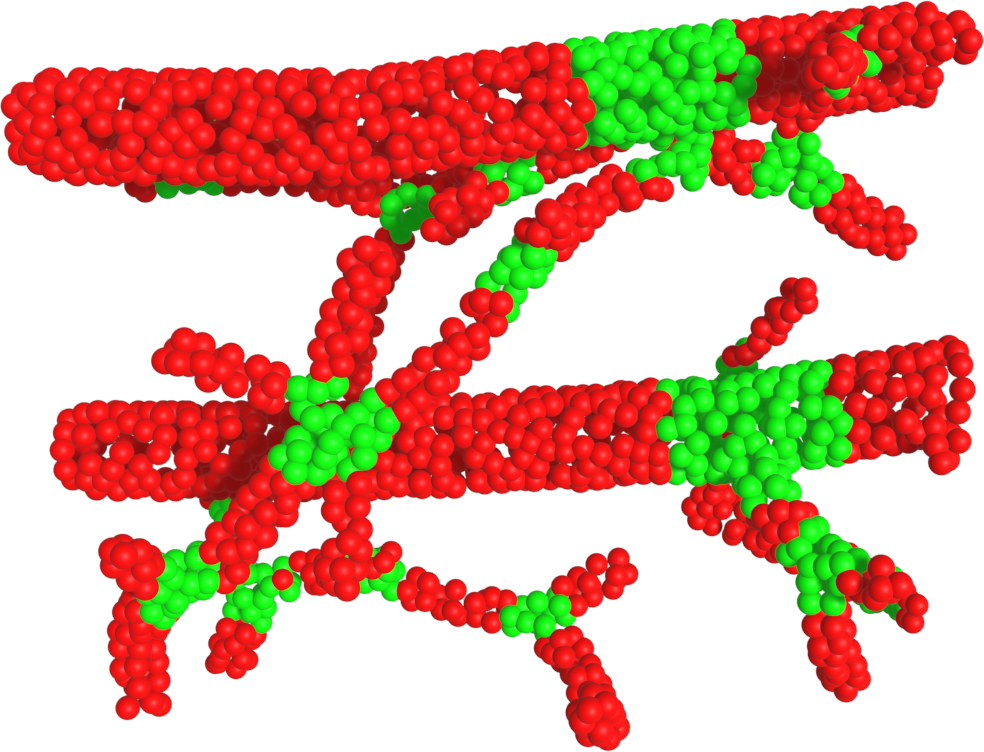} &
		\includegraphics[width=.135\linewidth]{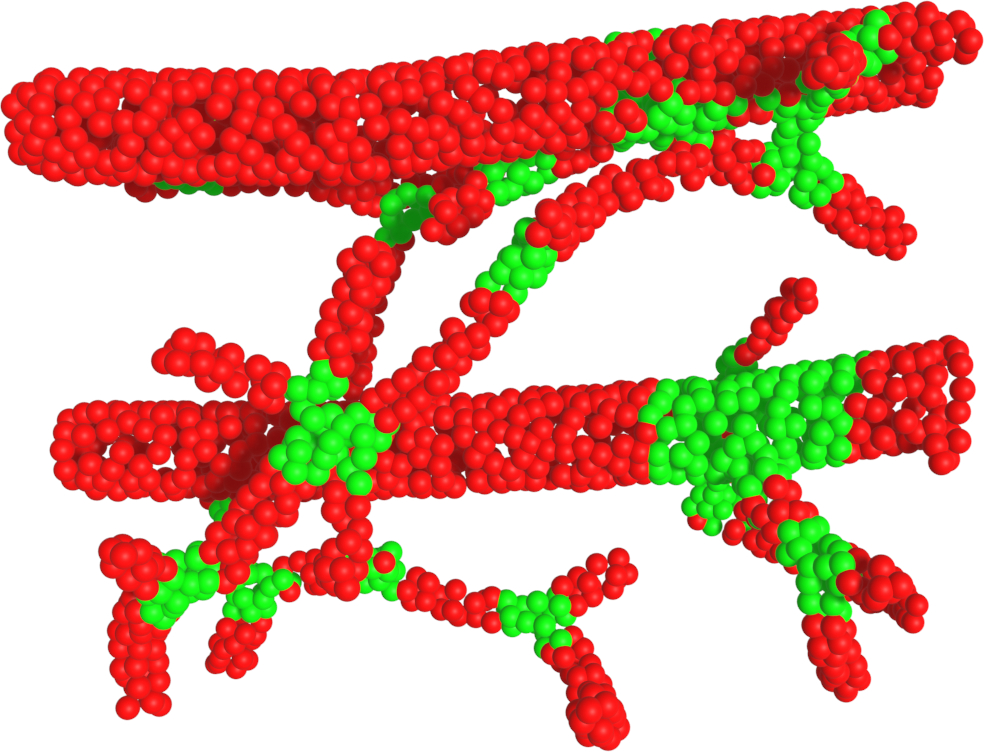} & 
		\includegraphics[width=.135\linewidth]{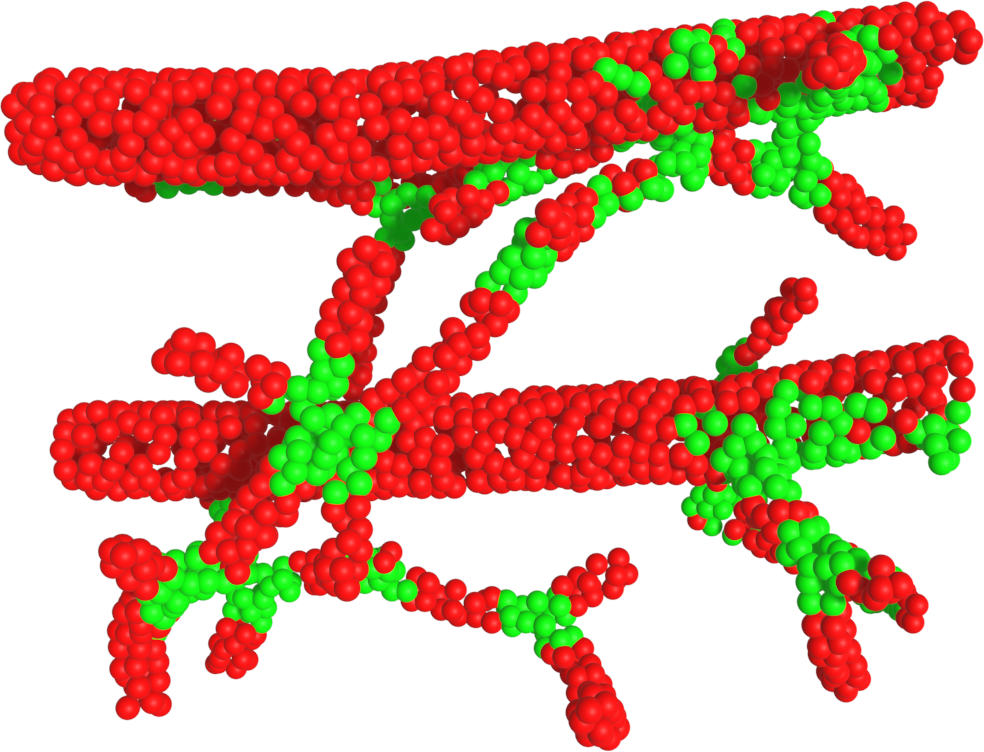} &  
		\includegraphics[width=.135\linewidth]{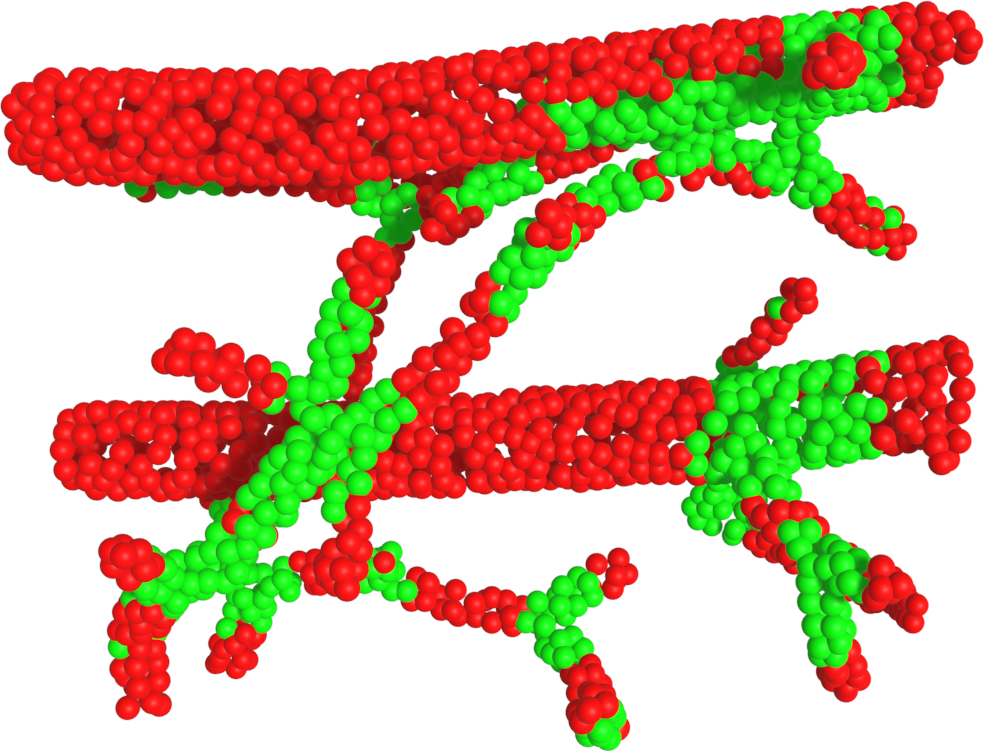} &  
		\includegraphics[width=.135\linewidth]{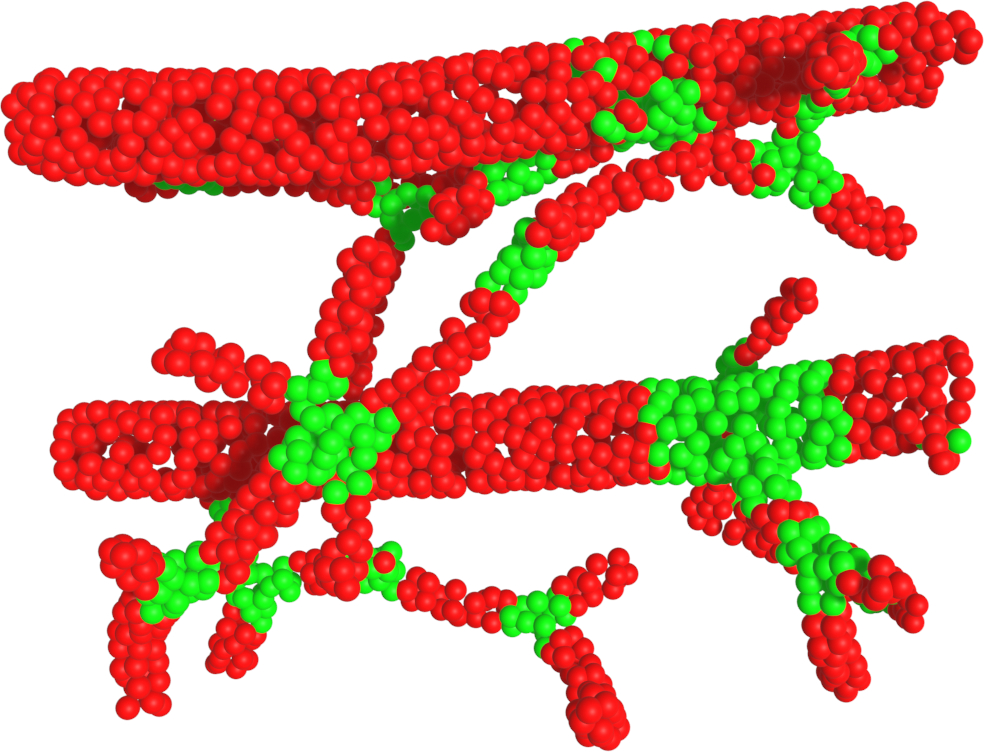} &  
		\includegraphics[width=.135\linewidth]{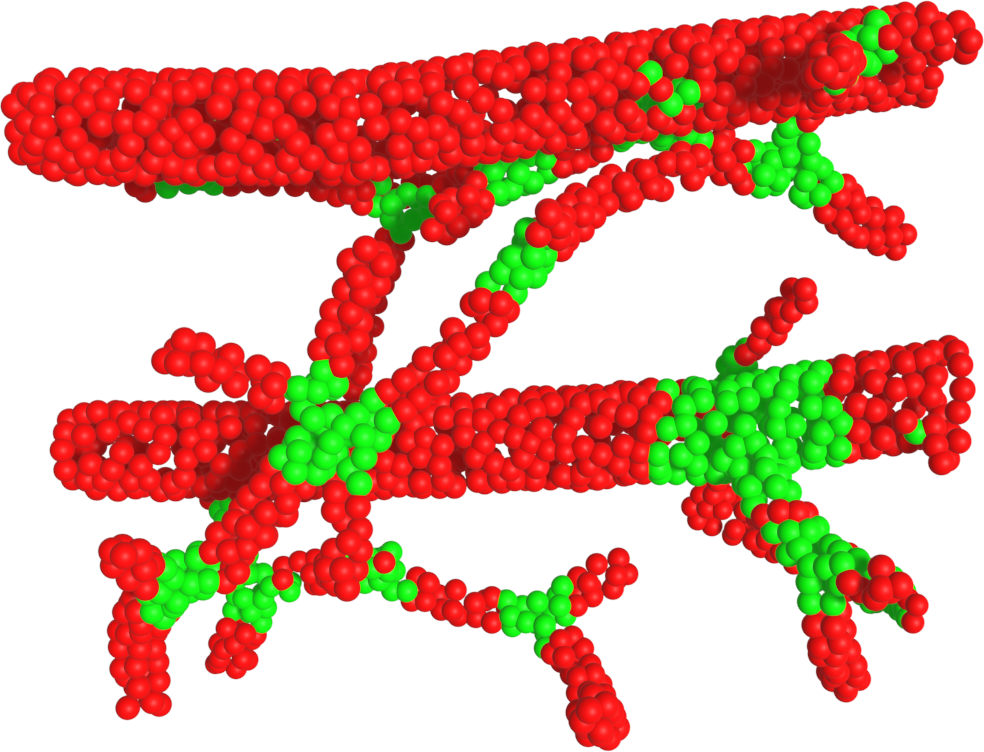} & 
		\includegraphics[width=.135\linewidth]{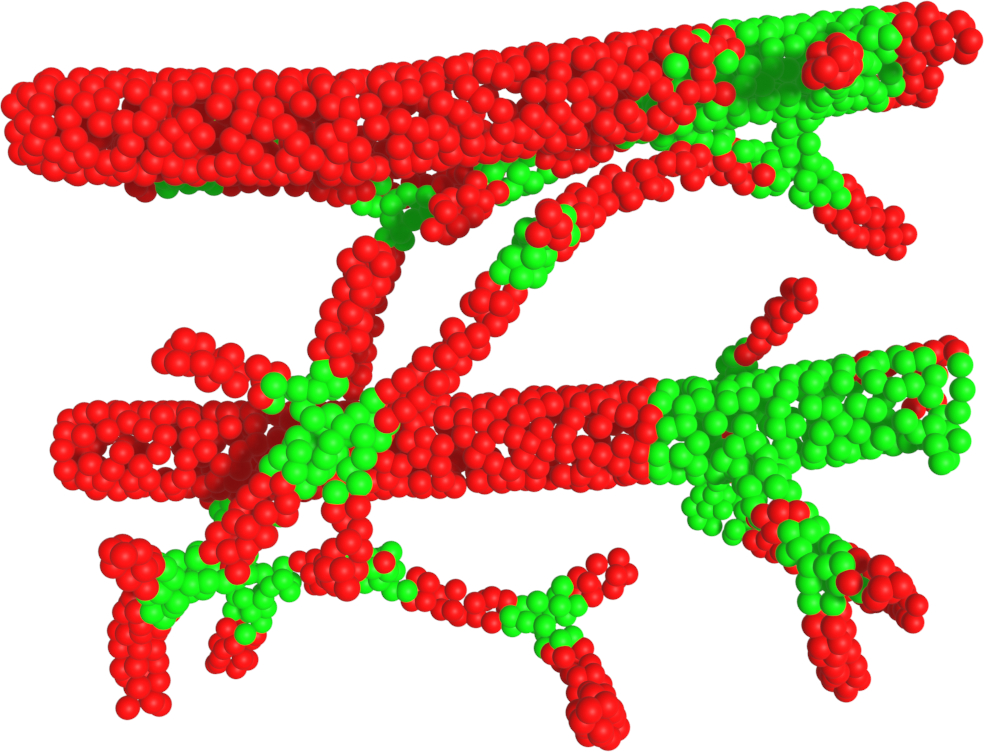} 
		%\vspace{-1mm}  
		\\ 
		\rotatebox[origin=l]{90}{\textit{PW$_{a}$ $\rightarrow$ PV}} &
		\includegraphics[width=.135\linewidth]{./figures/quali/gt_dvn.jpg} &
		\includegraphics[width=.135\linewidth]{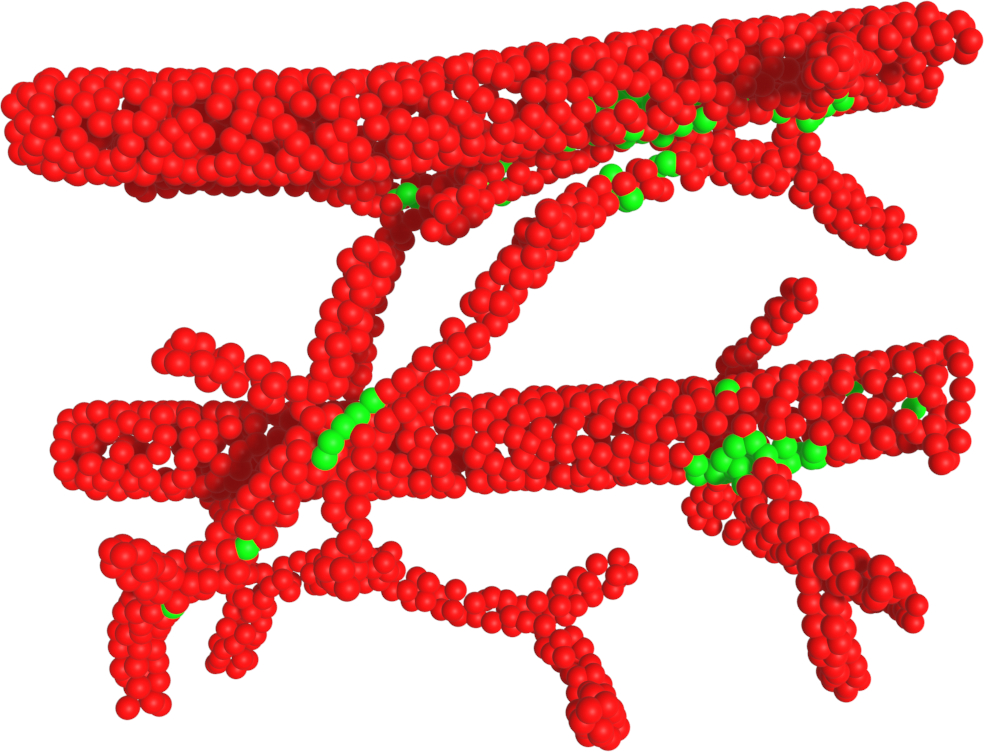} & 
		\includegraphics[width=.135\linewidth]{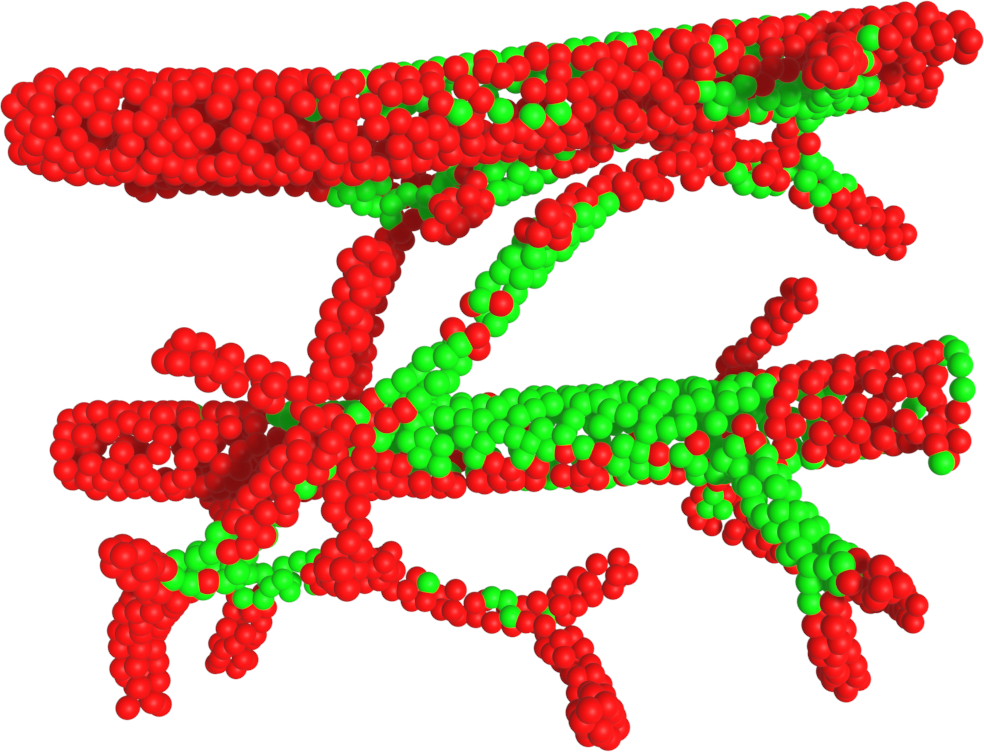} &  
		\includegraphics[width=.135\linewidth]{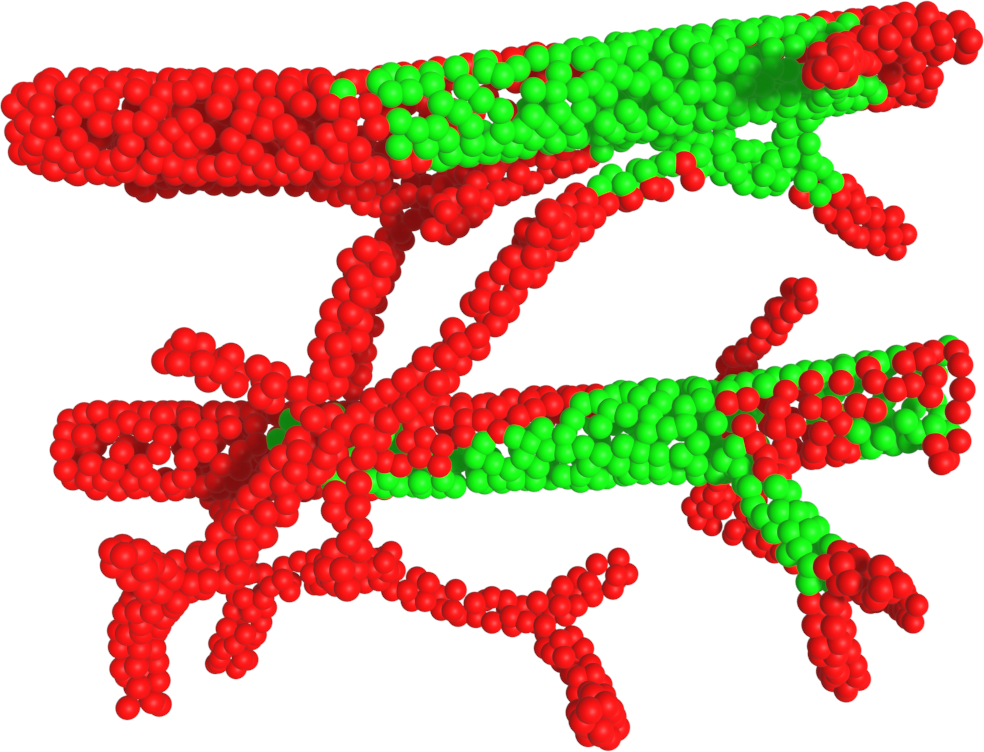} &  
		\includegraphics[width=.135\linewidth]{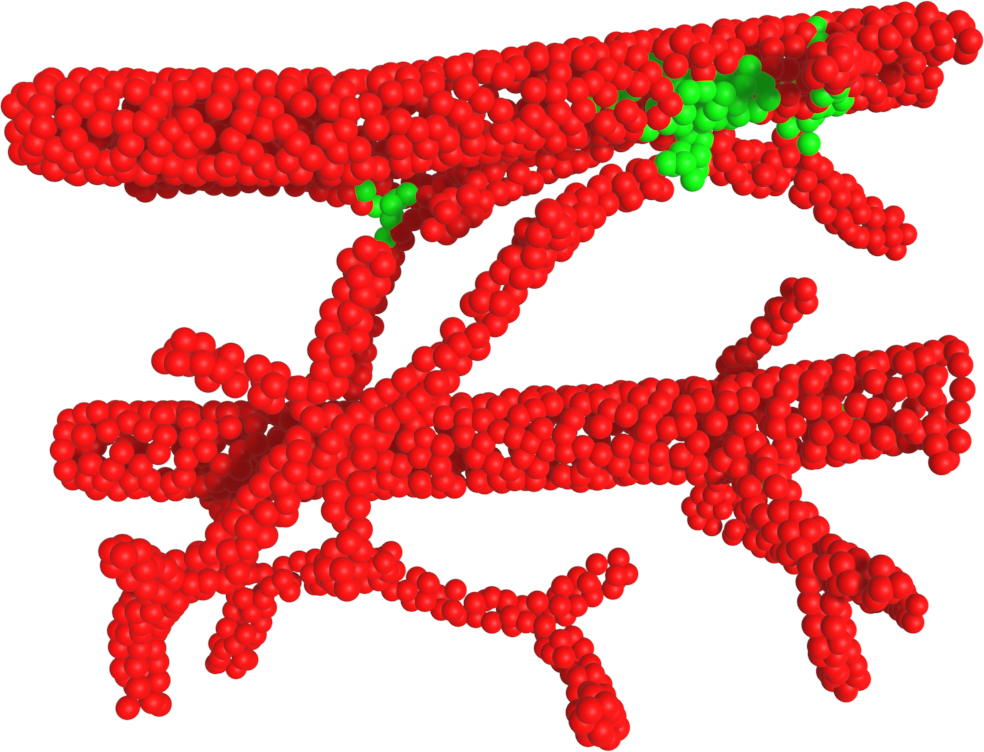} &  
		\includegraphics[width=.135\linewidth]{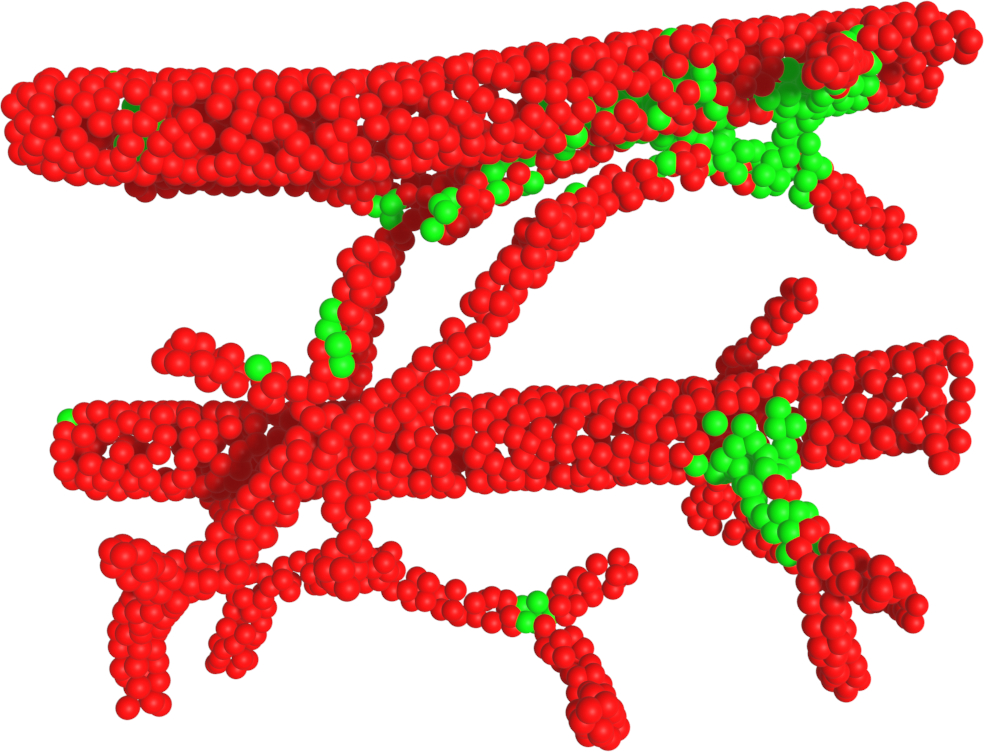} & 
		\includegraphics[width=.135\linewidth]{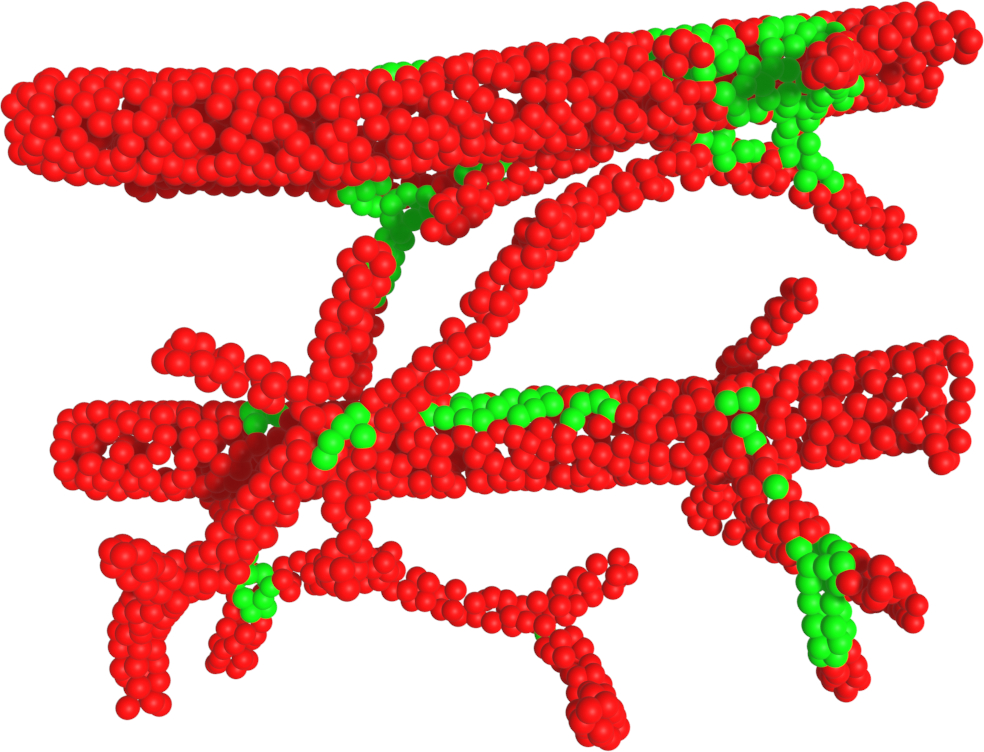} 
		%\vspace{-1mm}
		\\ 
		\rotatebox[origin=l]{90}{\textit{PW$_{b}$ $\rightarrow$ PV}} &
		\includegraphics[width=.135\linewidth]{./figures/quali/gt_dvn.jpg} &
		\includegraphics[width=.135\linewidth]{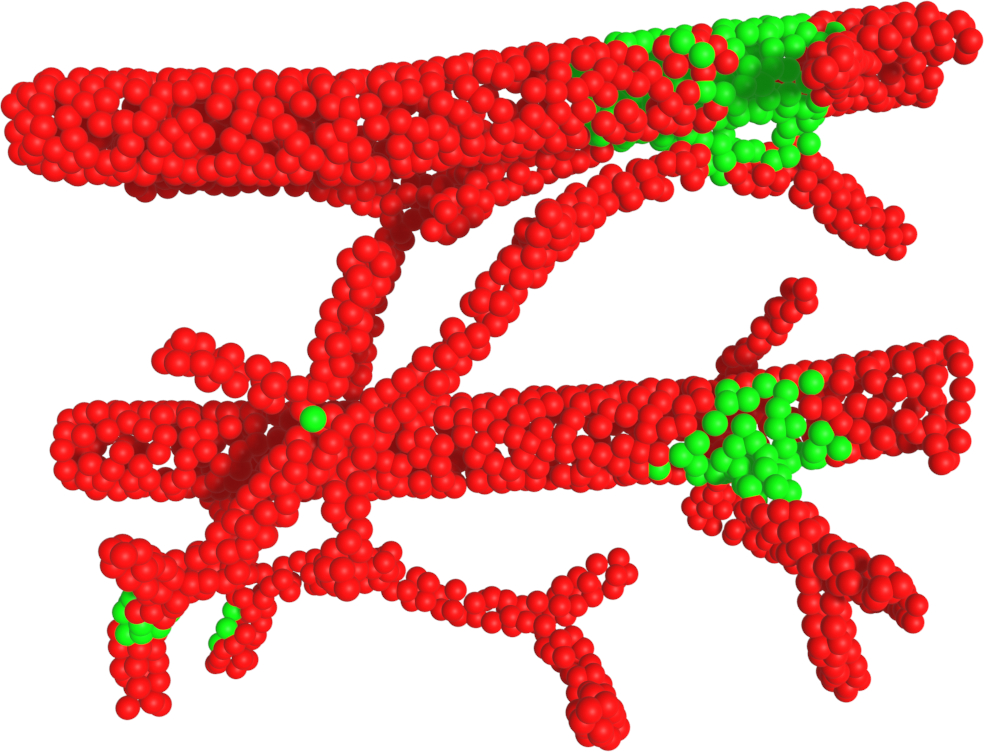} & 
		\includegraphics[width=.135\linewidth]{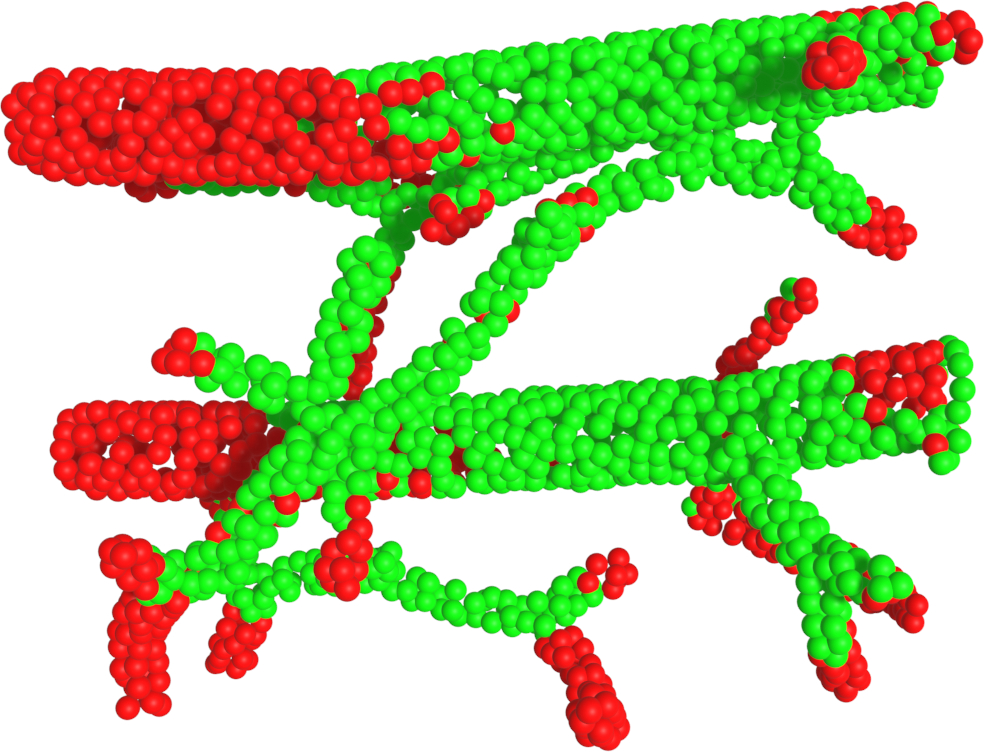} &  
		\includegraphics[width=.135\linewidth]{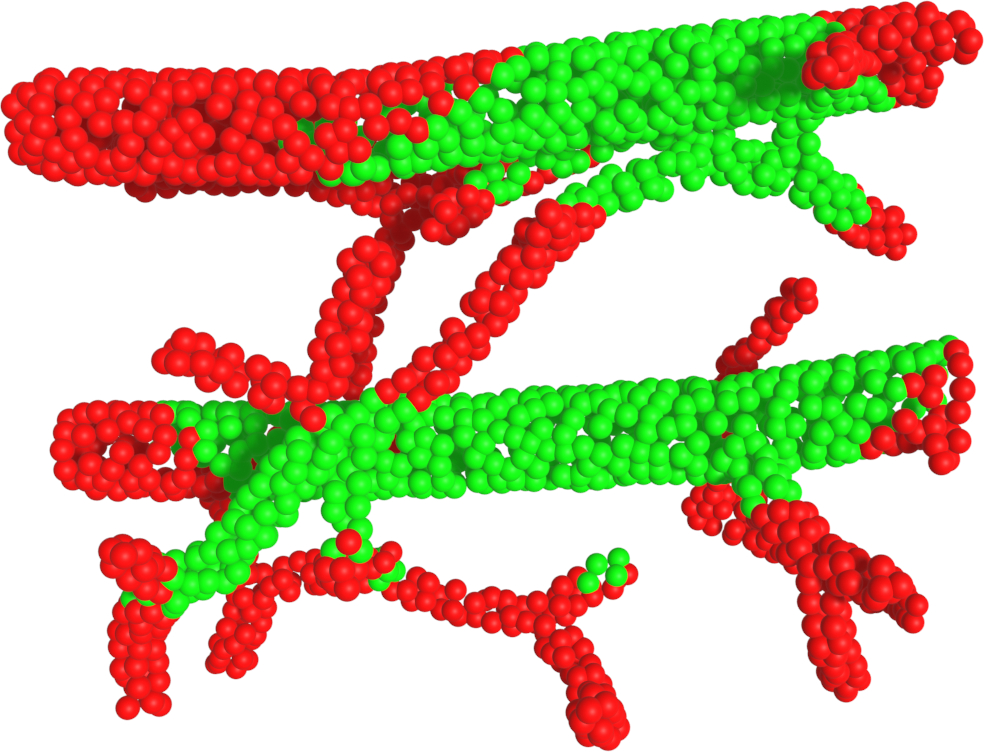} &  
		\includegraphics[width=.135\linewidth]{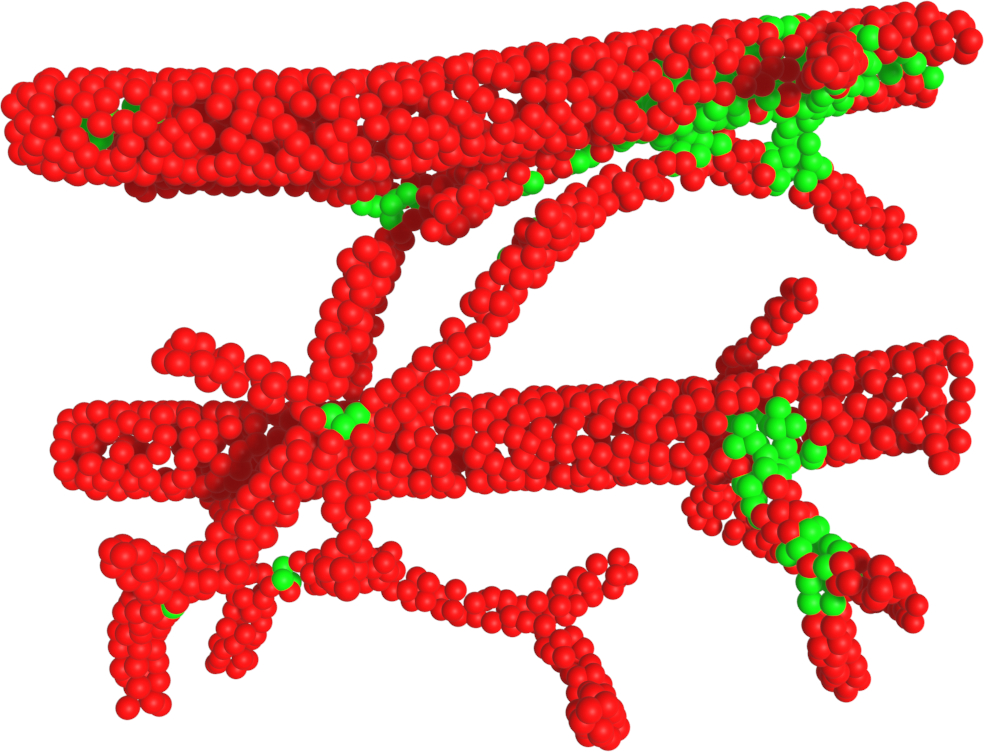} & 
		\includegraphics[width=.135\linewidth]{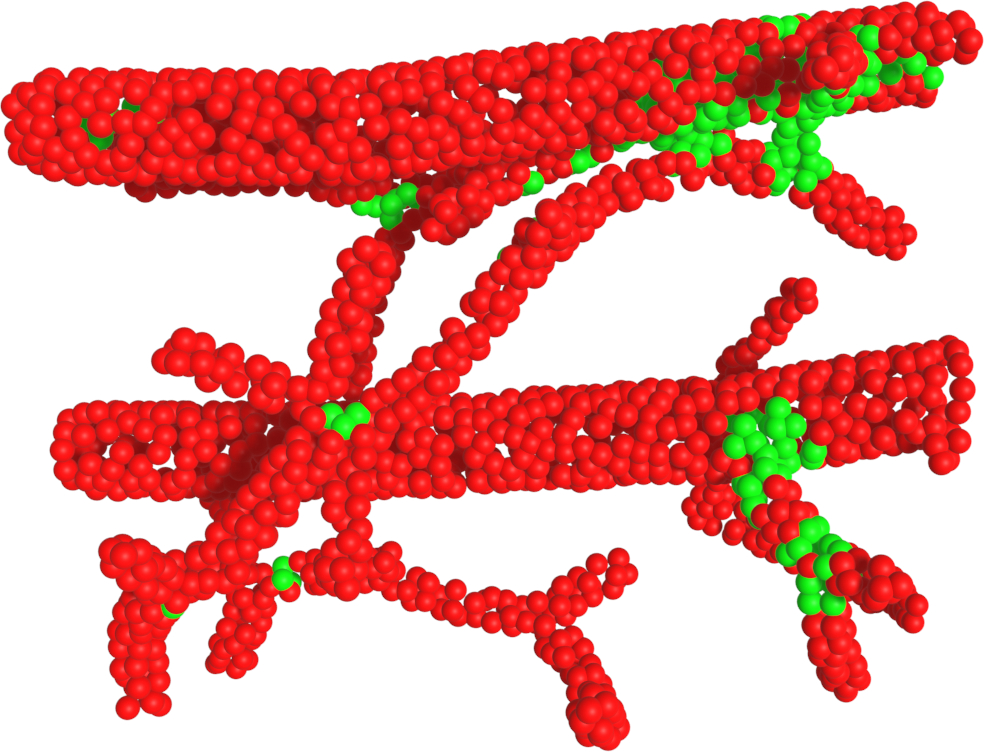} & 
		\includegraphics[width=.135\linewidth]{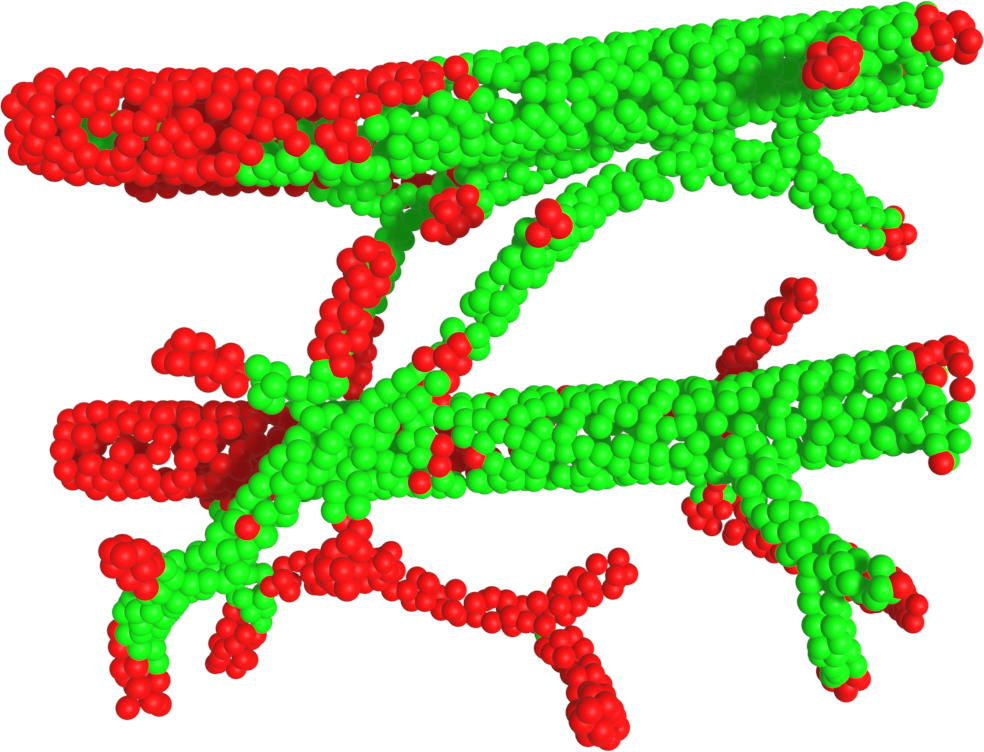} 
		%\vspace{-1mm} 
	\end{tabular} 
	\caption{Qualitative results on PointVessel and transferability benchmark on PointVessel. PointWire all classes to PointVessel (PW$_{a}$ $\rightarrow$ PV), PointWire bifurcation to PointVessel (PW$_{b}$ $\rightarrow$ PV). We show vessels and bifurcations in \textcolor{red}{red} and \textcolor{green}{green}.}
	\label{fig:transfer_quali}
    %\vspace{-5mm}
\end{figure*}

The transferability benchmark is intended to evaluate and demonstrate the selected point cloud segmentation methods on their ability to learn general concepts of DLOs topology. Moreover, these experiments serve as a more general evaluation of point learning methods as it gives more insight into the coupled performance of a network on how well it learns one specific domain and if it is capable of understanding high-level concepts that could be transferred to analogous domains.

Therefore, we train six point cloud segmentation methods \cite{qi2017pointnet++, wang2019dynamic, guo2021pct, xiang2021walk, Wiersma2022DeltaConv, ran2022surface} on the PointWire and the PointVessel datasets separately, and then cross-evaluated on the other's dataset test set. For example, we train the baseline methods on the PointVessel train set and evaluated them on the PointWire test set, and vice-versa.

\textbf{PointVessel $\rightarrow$ PointWire} The quantitative results of the transferability experiment by training first on the PointVessel train set and then evaluating on the PointWire test set are reported in Tab.~\ref{tab:tab3} in every even row. PointNet++ \cite{qi2017pointnet++} outperforms the other state-of-the-art methods and achieves a mIoU of $47.98~\%$. This demonstrates quite good generalization capabilities of PointNet++ when trained on complete synthetic point cloud data. Furthermore, a general observation is that training on complete synthetic PointVessel data leads to unsatisfied generalization on the more challenging incomplete, and noisy PointWire dataset for most of the other state-of-the-art methods. This is the case, especially for the bifurcation class. Most of the baseline methods achieve single-digit IoU results. The qualitative results shown in the third row of Fig.~\ref{fig:base_quali} confirm the aforementioned observations.

\textbf{PointWire $\rightarrow$ PointVessel} The quantitative results of the transferability experiments by training separately on the PointWire$_{all}$ and  PointWire$_{bifurcation}$ train sets and then evaluating on the PointVessel test set are reported in Tab.~\ref{tab:tab3}. Compared to the \textit{PointVessel $\rightarrow$ PointWire} benchmark, most of the models achieve relatively good performances, especially on the bifurcation class. DGCNN outperforms in terms of mIoU the second best performing method DeltaConv when trained on all PointWire classes by almost $6~\%$, and is on par when trained on the PointWire$_{bifurcation}$ train set. This indicates that point cloud graph representations learn well general topological DLO segmentation concepts. Moreover, the surface representation learning-based methods \cite{Wiersma2022DeltaConv, ran2022surface} are also reasonably good at capturing general DLO topology. The qualitative results are given in the second and third rows of Fig.~\ref{fig:transfer_quali}.

\subsection{Disentanglement Benchmark}

The quantitative evaluation results of the disentanglement benchmark are shown in Tab.~\ref{tab:tab3} in the last two columns. 
DGCNN yields the best results in terms of L-MAE, $0.79$, on the PointWire$_{bifurcation}$ dataset. This is very likely due to the heavy under-segmentation on thick DLO parts. DeltaConv performs best in terms of S-MAE on the PointWire$_{bifurcation}$ benchmark. This can be attributed to the best segmentation results.

Meanwhile, on the PointVessel disentanglement benchmark, the best performance in terms of L-MAE is achieved with PointNet++ and DeltaConv, which correlates also to their outperforming segmentation results.
Comparing the S-MAE results, we see that all the baselines are comparable.

\section{Conclusion}

We introduced two point cloud DLO datasets, i.e.~PointWire and PointVessel. The PointWire dataset is based on real wiring harness scans, while the PointVessel dataset contains synthetic blood vessel point clouds. We proposed a semi-automatic pipeline for generating point cloud-based wiring harness datasets. Furthermore, we conducted a transferability benchmark to analyze the generalization capabilities of state-of-the-art point cloud segmentation methods in terms of 3D DLO topology segmentation and disentanglement. We evaluated six point cloud segmentation methods and the results on the PointWire benchmark indicate that DLO topology segmentation is a hard challenge for current state-of-the-art methods. We strongly believe that the joint topology segmentation and disentanglement tasks can have a significant impact on the industry, and further research is encouraged.

% {\small
% \bibliographystyle{ieee_fullnam}
% \bibliography{ieeeconf/reference}
% }
% \bibliography{ieeeconf/reference}
\bibliographystyle{ieeetr}
\bibliography{ieeeconf/reference}

\begin{thebibliography}{10}

\bibitem{nguyen2021manufacturing}
H.~G. Nguyen, M.~Kuhn, and J.~Franke, ``Manufacturing automation for automotive wiring harnesses,'' {\em Procedia CIRP}, vol.~97, pp.~379--384, 2021.

\bibitem{navas2022wire}
G.~E. Navas-Reascos, D.~Romero, J.~Stahre, and A.~Caballero-Ruiz, ``Wire harness assembly process supported by collaborative robots: Literature review and call for r\&d,'' {\em Robotics}, vol.~11, no.~3, p.~65, 2022.

\bibitem{nguyen2021novel}
T.~P. Nguyen and J.~Yoon, ``A novel vision-based method for 3d profile extraction of wire harness in robotized assembly process,'' {\em Journal of Manufacturing Systems}, vol.~61, pp.~365--374, 2021.

\bibitem{kicki2021tell}
P.~Kicki, M.~Bednarek, P.~Lembicz, G.~Mierzwiak, A.~Szymko, M.~Kraft, and K.~Walas, ``Tell me, what do you see?—interpretable classification of wiring harness branches with deep neural networks,'' {\em Sensors}, vol.~21, no.~13, p.~4327, 2021.

\bibitem{caporali2022fastdlo}
A.~Caporali, K.~Galassi, R.~Zanella, and G.~Palli, ``Fastdlo: Fast deformable linear objects instance segmentation,'' {\em IEEE Robotics and Automation Letters}, vol.~7, no.~4, pp.~9075--9082, 2022.

\bibitem{valliani2019deep}
A.~A.-A. Valliani, D.~Ranti, and E.~K. Oermann, ``Deep learning and neurology: a systematic review,'' {\em Neurology and therapy}, vol.~8, pp.~351--365, 2019.

\bibitem{todorov2020machine}
M.~I. Todorov, J.~C. Paetzold, O.~Schoppe, G.~Tetteh, S.~Shit, V.~Efremov, K.~Todorov-V{\"o}lgyi, M.~D{\"u}ring, M.~Dichgans, M.~Piraud, {\em et~al.}, ``Machine learning analysis of whole mouse brain vasculature,'' {\em Nature methods}, vol.~17, no.~4, pp.~442--449, 2020.

\bibitem{kirst2020mapping}
C.~Kirst, S.~Skriabine, A.~Vieites-Prado, T.~Topilko, P.~Bertin, G.~Gerschenfeld, F.~Verny, P.~Topilko, N.~Michalski, M.~Tessier-Lavigne, {\em et~al.}, ``Mapping the fine-scale organization and plasticity of the brain vasculature,'' {\em Cell}, vol.~180, no.~4, pp.~780--795, 2020.

\bibitem{khandouzi2022retinal}
A.~Khandouzi, A.~Ariafar, Z.~Mashayekhpour, M.~Pazira, and Y.~Baleghi, ``Retinal vessel segmentation, a review of classic and deep methods,'' {\em Annals of Biomedical Engineering}, vol.~50, no.~10, pp.~1292--1314, 2022.

\bibitem{yang2020intra}
X.~Yang, D.~Xia, T.~Kin, and T.~Igarashi, ``Intra: 3d intracranial aneurysm dataset for deep learning,'' in {\em Proceedings of the IEEE/CVF Conference on Computer Vision and Pattern Recognition}, pp.~2656--2666, 2020.

\bibitem{liu2022edge}
Y.~Liu, J.~Liu, and Y.~Yuan, ``Edge-oriented point-cloud transformer for 3d intracranial aneurysm segmentation,'' in {\em Medical Image Computing and Computer Assisted Intervention--MICCAI 2022: 25th International Conference, Singapore, September 18--22, 2022, Proceedings, Part V}, pp.~97--106, Springer, 2022.

\bibitem{yang2023two}
X.~Yang, D.~Xia, T.~Kin, and T.~Igarashi, ``A two-step surface-based 3d deep learning pipeline for segmentation of intracranial aneurysms,'' {\em Computational Visual Media}, vol.~9, no.~1, pp.~57--69, 2023.

\bibitem{blender}
B.~O. Community, {\em Blender - a 3D modelling and rendering package}.
\newblock Blender Foundation, Stichting Blender Foundation, Amsterdam, 2018.

\bibitem{dvn}
G.~Tetteh, V.~Efremov, N.~D. Forkert, M.~Schneider, J.~Kirschke, B.~Weber, C.~Zimmer, M.~Piraud, and B.~H. Menze, ``Deepvesselnet: Vessel segmentation, centerline prediction, and bifurcation detection in 3-d angiographic volumes,'' {\em Frontiers in Neuroscience}, vol.~14, 2020.

\bibitem{modelnet}
Z.~Wu, S.~Song, A.~Khosla, F.~Yu, L.~Zhang, X.~Tang, and J.~Xiao, ``3d shapenets: A deep representation for volumetric shapes,'' in {\em Proceedings of the IEEE conference on computer vision and pattern recognition}, pp.~1912--1920, 2015.

\bibitem{chang2015shapenet}
A.~X. Chang, T.~Funkhouser, L.~Guibas, P.~Hanrahan, Q.~Huang, Z.~Li, S.~Savarese, M.~Savva, S.~Song, H.~Su, {\em et~al.}, ``Shapenet: An information-rich 3d model repository,'' {\em arXiv preprint arXiv:1512.03012}, 2015.

\bibitem{collins2022abo}
J.~Collins, S.~Goel, K.~Deng, A.~Luthra, L.~Xu, E.~Gundogdu, X.~Zhang, T.~F.~Y. Vicente, T.~Dideriksen, H.~Arora, {\em et~al.}, ``Abo: Dataset and benchmarks for real-world 3d object understanding,'' in {\em Proceedings of the IEEE/CVF Conference on Computer Vision and Pattern Recognition}, pp.~21126--21136, 2022.

\bibitem{koch2019abc}
S.~Koch, A.~Matveev, Z.~Jiang, F.~Williams, A.~Artemov, E.~Burnaev, M.~Alexa, D.~Zorin, and D.~Panozzo, ``Abc: A big cad model dataset for geometric deep learning,'' in {\em Proceedings of the IEEE/CVF conference on computer vision and pattern recognition}, pp.~9601--9611, 2019.

\bibitem{wu2021deepcad}
R.~Wu, C.~Xiao, and C.~Zheng, ``Deepcad: A deep generative network for computer-aided design models,'' in {\em Proceedings of the IEEE/CVF International Conference on Computer Vision}, pp.~6772--6782, 2021.

\bibitem{uy-scanobjectnn-iccv19}
M.~A. Uy, Q.-H. Pham, B.-S. Hua, D.~T. Nguyen, and S.-K. Yeung, ``Revisiting point cloud classification: A new benchmark dataset and classification model on real-world data,'' in {\em International Conference on Computer Vision (ICCV)}, 2019.

\bibitem{wu2023omniobject3d}
T.~Wu, J.~Zhang, X.~Fu, Y.~Wang, J.~Ren, L.~Pan, W.~Wu, L.~Yang, J.~Wang, C.~Qian, {\em et~al.}, ``Omniobject3d: Large-vocabulary 3d object dataset for realistic perception, reconstruction and generation,'' {\em arXiv preprint arXiv:2301.07525}, 2023.

\bibitem{laplace}
J.~Cao, A.~Tagliasacchi, M.~Olson, H.~Zhang, and Z.~Su, ``Point cloud skeletons via laplacian-based contraction,'' in {\em Proc. of IEEE Conf. on Shape Modeling and Applications}, 2015.

\bibitem{qi2017pointnet}
C.~R. Qi, H.~Su, K.~Mo, and L.~J. Guibas, ``Pointnet: Deep learning on point sets for 3d classification and segmentation,'' in {\em Proceedings of the IEEE conference on computer vision and pattern recognition}, pp.~652--660, 2017.

\bibitem{qi2017pointnet++}
C.~R. Qi, L.~Yi, H.~Su, and L.~J. Guibas, ``Pointnet++: Deep hierarchical feature learning on point sets in a metric space,'' {\em Advances in neural information processing systems}, vol.~30, 2017.

\bibitem{li2018pointcnn}
Y.~Li, R.~Bu, M.~Sun, W.~Wu, X.~Di, and B.~Chen, ``Pointcnn: Convolution on x-transformed points,'' {\em Advances in neural information processing systems}, vol.~31, 2018.

\bibitem{xu2018spidercnn}
Y.~Xu, T.~Fan, M.~Xu, L.~Zeng, and Y.~Qiao, ``Spidercnn: Deep learning on point sets with parameterized convolutional filters,'' in {\em Proceedings of the European conference on computer vision (ECCV)}, pp.~87--102, 2018.

\bibitem{lin2020fpconv}
Y.~Lin, Z.~Yan, H.~Huang, D.~Du, L.~Liu, S.~Cui, and X.~Han, ``Fpconv: Learning local flattening for point convolution,'' in {\em Proceedings of the IEEE/CVF conference on computer vision and pattern recognition}, pp.~4293--4302, 2020.

\bibitem{xu2021paconv}
M.~Xu, R.~Ding, H.~Zhao, and X.~Qi, ``Paconv: Position adaptive convolution with dynamic kernel assembling on point clouds,'' in {\em Proceedings of the IEEE/CVF Conference on Computer Vision and Pattern Recognition}, pp.~3173--3182, 2021.

\bibitem{wang2019dynamic}
Y.~Wang, Y.~Sun, Z.~Liu, S.~E. Sarma, M.~M. Bronstein, and J.~M. Solomon, ``Dynamic graph cnn for learning on point clouds,'' {\em Acm Transactions On Graphics (tog)}, vol.~38, no.~5, pp.~1--12, 2019.

\bibitem{zhao2021point}
H.~Zhao, L.~Jiang, J.~Jia, P.~H. Torr, and V.~Koltun, ``Point transformer,'' in {\em Proceedings of the IEEE/CVF international conference on computer vision}, pp.~16259--16268, 2021.

\bibitem{guo2021pct}
M.-H. Guo, J.-X. Cai, Z.-N. Liu, T.-J. Mu, R.~R. Martin, and S.-M. Hu, ``Pct: Point cloud transformer,'' {\em Computational Visual Media}, vol.~7, pp.~187--199, 2021.

\bibitem{xiang2021walk}
T.~Xiang, C.~Zhang, Y.~Song, J.~Yu, and W.~Cai, ``Walk in the cloud: Learning curves for point clouds shape analysis,'' in {\em Proceedings of the IEEE/CVF International Conference on Computer Vision}, pp.~915--924, 2021.

\bibitem{ran2022surface}
H.~Ran, J.~Liu, and C.~Wang, ``Surface representation for point clouds,'' in {\em Proceedings of the IEEE/CVF Conference on Computer Vision and Pattern Recognition}, pp.~18942--18952, 2022.

\bibitem{Wiersma2022DeltaConv}
R.~Wiersma, A.~Nasikun, E.~Eisemann, and K.~Hildebrandt, ``Deltaconv: Anisotropic operators for geometric deep learning on point clouds,'' {\em Transactions on Graphics}, vol.~41, July 2022.

\bibitem{mookiah2021review}
M.~R.~K. Mookiah, S.~Hogg, T.~J. MacGillivray, V.~Prathiba, R.~Pradeepa, V.~Mohan, R.~M. Anjana, A.~S. Doney, C.~N. Palmer, and E.~Trucco, ``A review of machine learning methods for retinal blood vessel segmentation and artery/vein classification,'' {\em Medical Image Analysis}, vol.~68, p.~101905, 2021.

\bibitem{fu2020rapid}
F.~Fu, J.~Wei, M.~Zhang, F.~Yu, Y.~Xiao, D.~Rong, Y.~Shan, Y.~Li, C.~Zhao, F.~Liao, {\em et~al.}, ``Rapid vessel segmentation and reconstruction of head and neck angiograms using 3d convolutional neural network,'' {\em Nature communications}, vol.~11, no.~1, p.~4829, 2020.

\bibitem{paetzold2021whole}
J.~C. Paetzold, J.~McGinnis, S.~Shit, I.~Ezhov, P.~B{\"u}schl, C.~Prabhakar, A.~Sekuboyina, M.~Todorov, G.~Kaissis, A.~Ert{\"u}rk, {\em et~al.}, ``Whole brain vessel graphs: A dataset and benchmark for graph learning and neuroscience,'' in {\em Thirty-fifth Conference on Neural Information Processing Systems Datasets and Benchmarks Track (Round 2)}, 2021.

\bibitem{caporali2022ariadne+}
A.~Caporali, R.~Zanella, D.~De~Greogrio, and G.~Palli, ``Ariadne+: Deep learning--based augmented framework for the instance segmentation of wires,'' {\em IEEE Transactions on Industrial Informatics}, vol.~18, no.~12, pp.~8607--8617, 2022.

\bibitem{choi2023mbest}
A.~Choi, D.~Tong, B.~Park, D.~Terzopoulos, J.~Joo, and M.~K. Jawed, ``mbest: Realtime deformable linear object detection through minimal bending energy skeleton pixel traversals,'' {\em arXiv preprint arXiv:2302.09444}, 2023.

\bibitem{rtdlo}
A.~Caporali, K.~Galassi, B.~L. Žagar, R.~Zanella, G.~Palli, and A.~C. Knoll, ``Rt-dlo: Real-time deformable linear objects instance segmentation,'' {\em IEEE Transactions on Industrial Informatics}, pp.~1--10, 2023.

\bibitem{lv2022learning}
K.~Lv, M.~Yu, Y.~Pu, and X.~Li, ``Learning to occlusion-robustly estimate 3-d states of deformable linear objects from single-frame point clouds,'' {\em arXiv preprint arXiv:2210.01433}, 2022.

\bibitem{nguyen2022enabling}
H.~G. Nguyen, R.~Habiboglu, and J.~Franke, ``Enabling deep learning using synthetic data: A case study for the automotive wiring harness manufacturing,'' {\em Procedia CIRP}, vol.~107, pp.~1263--1268, 2022.

\bibitem{cop2021new}
K.~P. Cop, A.~Peters, B.~L. {\v{Z}}agar, D.~Hettegger, and A.~C. Knoll, ``New metrics for industrial depth sensors evaluation for precise robotic applications,'' in {\em 2021 IEEE/RSJ International Conference on Intelligent Robots and Systems (IROS)}, pp.~5350--5356, IEEE, 2021.

\bibitem{zhou2018open3d}
Q.-Y. Zhou, J.~Park, and V.~Koltun, ``Open3d: A modern library for 3d data processing,'' {\em arXiv preprint arXiv:1801.09847}, 2018.

\bibitem{VTK4}
W.~Schroeder, K.~Martin, and B.~Lorensen, {\em {The Visualization Toolkit--An Object-Oriented Approach To 3D Graphics}}.
\newblock Kitware, Inc., fourth~ed., 2006.

\bibitem{li2022vbnet}
Y.~Li, T.~Ren, J.~Li, H.~Wang, X.~Li, and A.~Li, ``Vbnet: An end-to-end 3d neural network for vessel bifurcation point detection in mesoscopic brain images,'' {\em Computer Methods and Programs in Biomedicine}, vol.~214, p.~106567, 2022.

\bibitem{meyer2023cherrypicker}
L.~Meyer, A.~Gilson, O.~Scholz, and M.~Stamminger, ``Cherrypicker: Semantic skeletonization and topological reconstruction of cherry trees,'' in {\em Proceedings of the IEEE/CVF Conference on Computer Vision and Pattern Recognition}, pp.~6243--6252, 2023.

\bibitem{goyal2017accurate}
P.~Goyal, P.~Doll{\'a}r, R.~Girshick, P.~Noordhuis, L.~Wesolowski, A.~Kyrola, A.~Tulloch, Y.~Jia, and K.~He, ``Accurate, large minibatch sgd: Training imagenet in 1 hour,'' {\em arXiv preprint arXiv:1706.02677}, 2017.

\bibitem{qian2022pointnext}
G.~Qian, Y.~Li, H.~Peng, J.~Mai, H.~A. A.~K. Hammoud, M.~Elhoseiny, and B.~Ghanem, ``Pointnext: Revisiting pointnet++ with improved training and scaling strategies,'' {\em arXiv preprint arXiv:2206.04670}, 2022.

\end{thebibliography}

\end{document}